\documentclass[journal]{IEEEtai}

\usepackage[colorlinks,urlcolor=blue,linkcolor=blue,citecolor=blue]{hyperref}

\usepackage{array}
\usepackage{graphicx}
\usepackage{amssymb}
\usepackage{amsmath}
\usepackage{booktabs}
\usepackage{multirow}  
\usepackage{algorithm}
\usepackage{algorithmic}
\usepackage{subcaption}
\usepackage{xcolor}

\usepackage{pifont}
\usepackage{colortbl}
\usepackage{makecell}
\usepackage{tabularx}
\usepackage{ragged2e}
\usepackage{siunitx}
\usepackage{cuted}
\usepackage{pdfpages}
\usepackage{tikzpagenodes}

\newenvironment{myquote}[1][2em]
  {\list{}{\leftmargin=#1 \rightmargin=#1}
   \item\relax}
  {\endlist}
\definecolor{MyGreen}{RGB}{224,245,241}
\definecolor{UpRed}{RGB}{220,48,150}
\definecolor{DownGreen}{RGB}{107,146,96}

\begin{document}

\title{NL2GenSym: Natural Language to Generative Symbolic Rules for SOAR Cognitive Architecture via Large Language Models} 

\author{Fang Yuan, Junjie Zeng, Yue Hu, Zhengqiu Zhu, Quanjun Yin, Yuxiang Xie
\thanks{This work was supported in part by the National Natural Science Foundation of China (Grant No. 62306329), the Hunan Provincial Natural Science Foundation of China (Grant No. 2023JJ40676), the China Association for Science and Technology (CAST) Youth Talent Supporting Program (Grant No. 2024-JCJQ-QT-034), and Independent Innovation Foundation of National University of Defense Technology (Grant No. 24-ZZCX-JDZ-50). \textit{(Corresponding author: Junjie Zeng)}}
\thanks{Fang Yuan is with the College of Systems Engineering, National University of Defense Technology, and also with the State Key Laboratory of Digital Intelligent Modeling and Simulation, Changsha 410073, China, and also with Test Center, National University of Defense Technology, Xi'an 710106, China (e-mail: fangyuan17@nudt.edu.cn).}
\thanks{Junjie Zeng, Yue Hu, Zhengqiu Zhu, Quanjun Yin are with the College of Systems Engineering, National University of Defense Technology, Changsha 410073, China (e-mail: \{zengjunjie13, huyue11, zhuzhengqiu12\}@nudt.edu.cn, yin\_quanjun@163.com).}
\thanks{Yuxiang Xie is with the College of Systems Engineering, National University of Defense Technology, Changsha 410073, China (e-mail: yxxie@nudt.edu.cn).}
}
\markboth{Journal of IEEE Transactions on Artificial Intelligence, Vol. 00, No. 0, Month 2020}
{ \MakeLowercase{Yuan \textit{et al.}}: NL2GenSym: Natural Language to Generative Symbolic Rules}

\maketitle

\begin{abstract}
SOAR, a classic symbol-based cognitive architecture, has been fostering the development of general, human-like intelligent agents. Nevertheless, its practical adoption is hindered by the laborious manual rule coding. Emerging Large Language Models (LLMs) present the immense potential for efficient rules generation. However, there is a critical gap that current research predominantly focuses on conceptual frameworks and lacks robust experimental validation.
To bridge this gap, we propose \textit{N}atural \textit{L}anguage to \textit{Gen}erative \textit{Sym}bolic Rules (NL2GenSym), a novel framework that integrates LLMs with SOAR to autonomously produce generative symbolic rules from natural language. 
Specifically, our framework introduces a novel Execution-Grounded Generator-Critic mechanism. The LLM-based Generator, guided by a Retrieval-Augmented Generation-accessed self-evolving domain knowledge base, proposes rules from natural language. Subsequently, these rules are immediately executed within the SOAR environment to rigorously validate their correctness. Based on this execution-grounded feedback, a reflective LLM-based Critic drives the iterative refinement of these rules.
Experiments on our specialized Water Jug Problem (WJP) dataset, utilizing both Gemini and Qwen series models, validate the efficacy of our framework. It achieves a success rate over 86\% in generating rules from natural language. Crucially, the framework also generates novel heuristic rules, reducing average decision cycles for solving the WJP to 1.98 times the optimal solution and 1/1000 of baseline methods. Additionally, our initial experiments show that NL2GenSym enables smaller-parameter models to achieve better performance than larger counterparts.
\end{abstract}

\begin{IEEEImpStatement}
To our knowledge, NL2GenSym is the first framework to achieve a successful end-to-end integration of LLMs with SOAR. It comprehensively covers the entire pipeline, from framework design to robust experimental implementation and validation.
Through its novel Execution-Grounded Generator-Critic mechanism, the framework not only translates natural language directly into executable symbolic rules but also generates novel heuristic rules for optimized problem-solving. 
By replacing the laborious and expertise-intensive process of manual rule coding, this innovation dramatically lowers the barrier to utilizing SOAR. 
This unlocks the potential to develop adaptive autonomous agents for complex domains requiring intricate symbolic reasoning, such as complex mission planning, target identification and allocation, and robotic control.
Critically, our work demonstrates that a well-designed architecture can be more impactful than sheer model scale, offering an efficient path toward artificial general intelligence.
\end{IEEEImpStatement}


\section{Introduction}
\label{introduction}

\IEEEPARstart{T}{he} pursuit of artificial general intelligence (AGI) has long been a central goal in Artificial Intelligence (AI) research.
Cognitive Architectures (CAs) have emerged as foundational tools for this endeavor~\cite{newell1994unified, laird2012soar}.
By emulating human cognitive processes((e.g., learning, reasoning), CAs provide unified frameworks for developing general intelligent agents~\cite{wooldridge1994agent, rosenbloom1985r1}.
As exemplified by SOAR~\cite{kotseruba202040}, their integration of structured knowledge representation, explicit reasoning, and adaptive learning (e.g., chunking~\cite{laird1986chunking}) have been proven advantageous in complex problem-solving, cognitive simulation, and agent behavior generation~\cite{muni2023better, ramzani2024recognition}.
However, CAs including SOAR require rules to be coded in their specialized symbolic languages, which poses a significant hurdle for users. Traditional manual coding process is not only time-consuming and labor-intensive but also demands extensive experience and deep domain knowledge. Furthermore, It becomes even less feasible when tackling novel tasks for which no prior knowledge or expertise exists. 

In recent years, the rapid development of Large Language Models (LLMs) has brought revolutionary breakthroughs to the field of AI~\cite{horawalavithana2023scitune, hagos2024recent, adornetto2025generative}.
Pre-trained on massive text corpora, LLMs have acquired powerful natural language understanding, generation and learning capabilities, demonstrating impressive performance on a variety of tasks~\cite{valmeekam2023planbench, xie2024travelplanner, pallagani2024prospects}.

Consequently, LLMs offer the potential to efficiently generate symbolic rules directly through natural language interaction. Researchers have begun to explore leveraging LLMs to empower SOAR, aiming to construct more robust and general AI systems. 
For instance, Laird et al.~\cite{laird2023proposal} explored the potential integration of Transformers as a learning and memory component of SOAR. Kirk et al.~\cite{kirk2023exploiting} identified three conceptual integration patterns for using LLMs as knowledge sources in cognitive systems. Wu et al.~\cite{wu2023comparing} presented the design and development process for an ACT-R/SOAR model utilizing LLMs as interactive interfaces. While these works predominantly concentrate on conceptual framework development and integration pattern proposals, often accompanied by only preliminary experiments. 

At present, robust experimental validation is largely absent, and as a result, the field has yet to see established, impactful results from LLM-assisted cognitive architecture modeling. This presents the following key challenges for realizing the potential of LLM-empowered SOAR:
\begin{enumerate}
\item[1)] Existing theoretical frameworks for LLM-SOAR integration lack robust experimental validation, as the clash between the probabilistic nature of LLMs and the deterministic syntax of SOAR hinders their practical viability. The first challenge is therefore how to successfully transition from the traditional manual coding approach towards the automated generation of executable symbolic rules from natural language.
\item[2)] While generating executable rules addresses the initial barrier of entry, it does not guarantee their efficiency. The second challenge is therefore how to facilitate continuous optimization of generated rules and the emergence of novel, high-performance rules.
\end{enumerate}

To tackle these challenges, we propose \textit{N}atural \textit{L}anguage to \textit{Gen}erative \textit{Sym}bolic Rules (NL2GenSym), a novel framework that effectively integrates LLMs with SOAR to enable the automated generation and continuous optimization of executable symbolic rules from natural language.
NL2GenSym operates through a closed-loop, Execution-Grounded Generator-Critic mechanism, comprising an LLM-based Generator and a reflective LLM-based Critic. 
Both modules leverage a self-evolving domain knowledge base via Retrieval-Augmented Generation (RAG)~\cite{lewis2020retrieval, Supparesk2025Chatbot}. 
We construct this knowledge base to provide essential SOAR syntax, semantics, and a dynamically updated cases pool enriched by optimal ``seed" rules from SOAR execution. 
The Generator first translates natural language into symbolic rules, which are immediately executed in the SOAR environment to provide execution-grounded feedback. The Critic then analyzes this feedback to generate targeted guidance, enabling iterative rule refinement in subsequent cycles.

Our work provides a tangible and viable solution to effectively integrate LLMs with SOAR. The technical contributions of this work can be summarized as follows:

\begin{enumerate}
\item[1)] \textbf{Execution-Grounded, Closed-Loop Framework for Integrating LLMs with SOAR.} We introduce a novel framework that automates the generation of executable SOAR rules directly from natural language descriptions. This is achieved through our core execution-grounded generator–critic mechanism. By doing so, our framework fundamentally lowers the barrier to using SOAR, empowering users without SOAR-specific expertise to effectively utilize the system.

\item[2)] \textbf{Self-Evolving Domain Knowledge Base and Reflective Critic Module.} To operationalize our framework, we design and implement two core technical components:
(1) The Domain Knowledge Base is comprised of two critical elements. The first is a repository of foundational knowledge, including essential SOAR syntax, semantics and examples, which we have specifically structured to ensure the LLM adheres to SOAR; The second is the dynamic case pool. This pool automatically accumulates optimal ``seed" rules from successful execution cycles, providing the Generator with a continuously improving repository of high-quality exemplars. (2) The Critic Module analyzes detailed execution traces from SOAR to generate structured, interpretable natural language feedback. This feedback is crucial for guiding the Generator toward effective and efficient rule refinement.

\item[3)] \textbf{Robust Experimental Validation and Emergent Capabilities.} We provide the first robust, end-to-end experimental validation of an LLM-SOAR integration. Our results demonstrate that NL2GenSym not only achieves a high success rate ($>$86\%) in generating functionally correct rules but, more importantly, facilitates the emergent discovery of novel, high-efficiency heuristic rules. This leads to a significant performance improvement: solving steps are reduced to near-optimal levels (1.98× the theoretical optimum), while outperforming baseline methods by a factor of 1000. Furthermore, the framework enables smaller-parameter models to surpass their larger counterparts, highlighting that well-designed architectures can outweigh sheer model scale.
\end{enumerate}

The remaining sections of this paper are outlined as follows: Section~\ref{sec:background} provides theoretical background. Section ~\ref{method} introduces our proposed NL2GenSym framework in detail. Section~\ref{experiments} presents a comprehensive analysis of the experimental results. Subsequently, Section~\ref{discussion} discusses the limitations of our current approach, Section~\ref{work} reviews related work in the field. Finally, in Section~\ref{conclusion}, we conclude this paper and discuss directions for future work.

\section{Theoretical background}
\label{sec:background}

\noindent \textbf{Problem Space Computational Model (PSCM)}
It is a general theory of decision-making and problem-solving proposed by Newell et al.~\cite{newell1993formulating}. The core idea of this model is that an agent's behavior manifests as a series of decision-making processes, utilizing its available knowledge to select appropriate actions (i.e., ``operators") to transition from an initial state to a goal state within a ``problem space". The SOAR cognitive architecture is built upon the principles of PSCM, providing a concrete computational implementation for this theoretical framework~\cite{newell1994unified, laird2012soar}. PSCM emphasizes understanding and constructing intelligent behavior through states, operators, and search, where knowledge is used to guide operator selection and the construction and navigation of the problem space.

\begin{figure}[htbp]
    \centering
    \captionsetup{justification=raggedright, singlelinecheck=false}
    \includegraphics[width=0.45\textwidth]{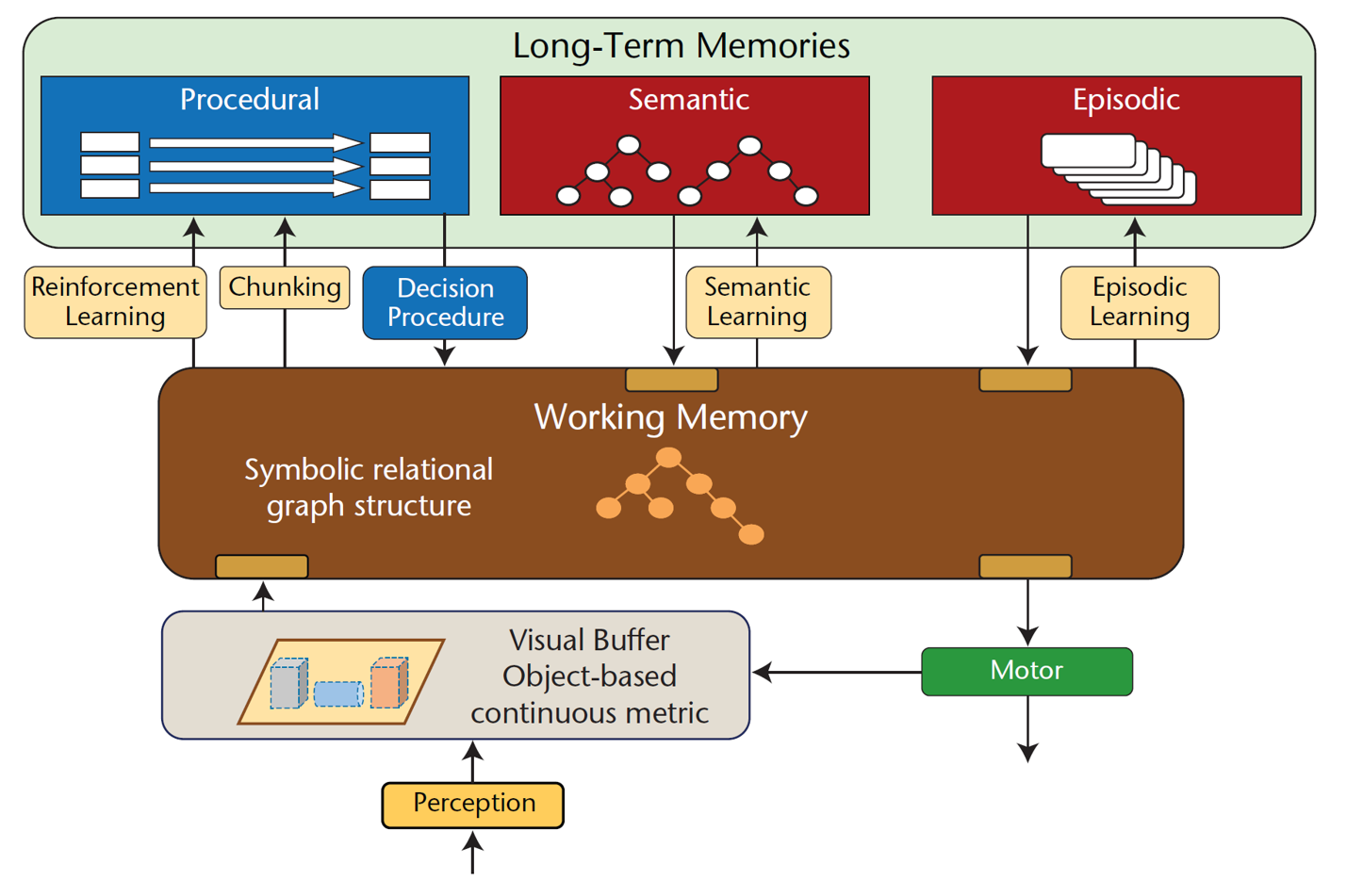}
    \caption{Basic architecture of the SOAR cognitive model~\cite{laird2012soar}.}
    \label{fig:soar_architecture}
\end{figure}

\noindent \textbf{SOAR Cognitive Architecture}
SOAR delineates the relationships between states, operators, and results. Conceived as an architecture for ``universal intelligence", the principal research focus is on exploring knowledge, cognition, intelligence, and memory as they manifest in human cognitive processes~\cite{newell1994unified}. SOAR supports such intelligent behavior through a series of fixed computational mechanisms and structures, with its basic composition typically including interacting memory modules and computational processes~\cite{laird2012soar}, as illustrated in Figure~\ref{fig:soar_architecture}.
 
Figure~\ref{fig:soar_architecture} shows the structure of SOAR, which consists of interacting task-independent modules. Working Memory (WM) maintains an agent’s situational awareness, including perceptions, goals, reasoning states and so on. Long-Term knowledge is stored in three memories. These include procedural memory (``if-then” knowledge); semantic memory (facts about the world and the agent); and episodic memory (memories of experiences). Procedural knowledge drives behavior by responding to the contents of WM and making modifications to it. Automatic learning mechanisms are associated with procedural and episodic memories~\cite{laird2012soar, laird2022analysis}.

The solving process of SOAR entails a continuous cycle of selecting operators, applying operators, and changing states until the desired goal is attained. This iterative procedure is commonly referred to as the ``decision cycle"~\cite{zhou2025hybrid}. It solves problems through the activation of production rules, offering a foundation for interpretable, human-like decision-making.

When SOAR encounters insufficient knowledge during its decision-making process (e.g., inability to select a unique operator, or unresolvable conflicts between multiple operators), an ``impasse" arises~\cite{laird2022introduction}. At this point, SOAR automatically creates a substate or subgoal, and within this substate, it employs the same decision cycle to attempt to resolve the impasse. This recursive, goal-driven problem-solving mechanism is key to the ability of SOAR to exhibit human-like deliberative behavior. Theoretically, this nesting of substates can proceed indefinitely, greatly enriching cognitive capabilities.

\begin{figure}[htbp]
  \centering
  \captionsetup{justification=justified,singlelinecheck=false}
  \includegraphics[width=0.7\textwidth, height=7cm, keepaspectratio]{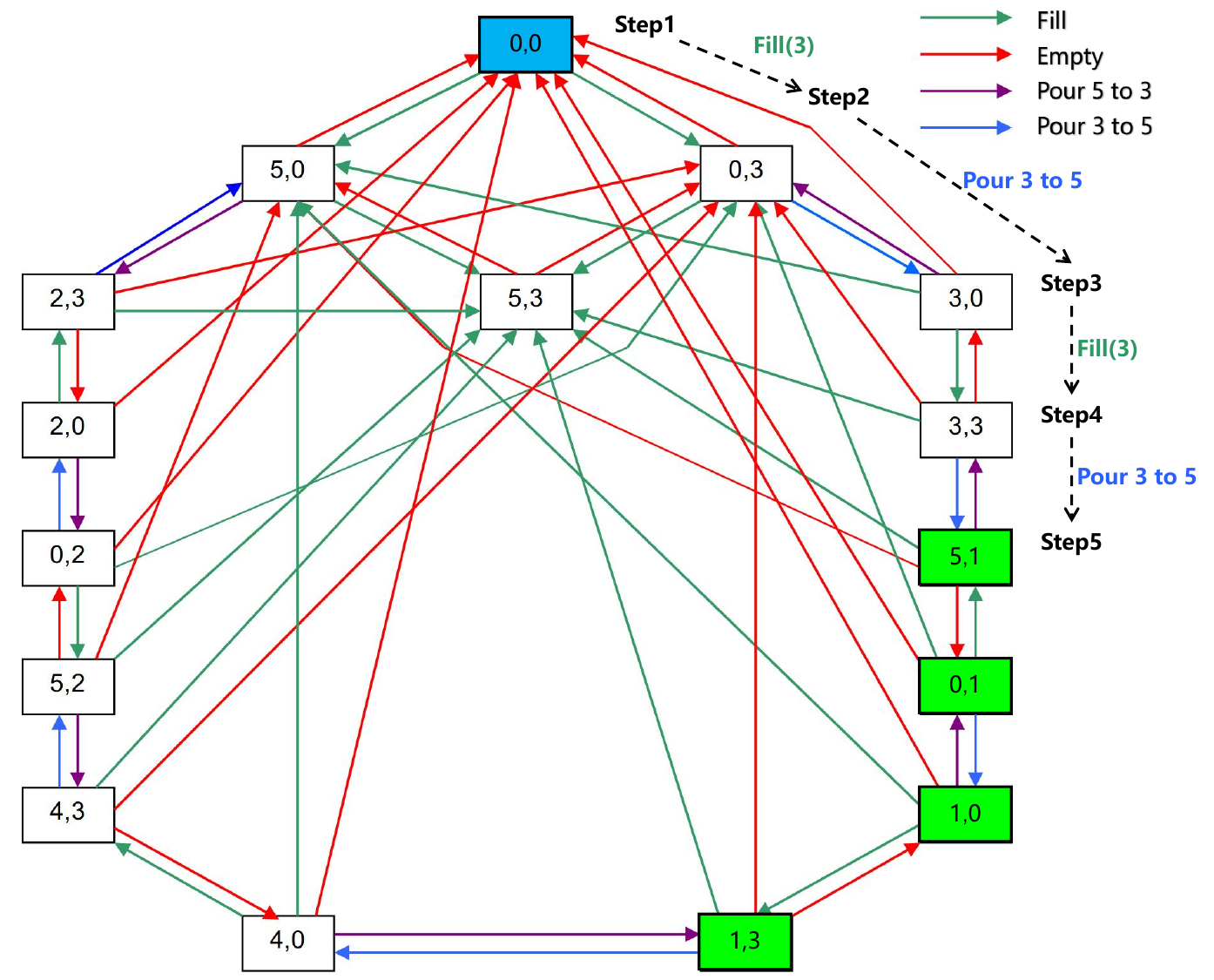}
  \caption{Problem space representation of the WJP. Each square \texttt{(x, y)} represents a state, corresponding to the water volume in the 5\,\text{L} and 3\,\text{L} jugs, respectively. The initial state \texttt{(0,0)} is denoted by the blue square, and a target goal state (any state containing 1\,\text{L}) is indicated by the green square. The colored arrows depict the state transitions performed by distinct Operators. These Operators include Fill  (jug) (to fill a jug to its volume), Empty (jug) (to empty its contents), and Pour (jug\_from, jug\_to) (to pour water between jugs until the source is empty or the destination is full). The path highlighted by dashed arrows illustrates the optimum solution, corresponding to five decision cycles in SOAR.}
  \label{waterjug}
\end{figure}

\begin{figure*}[htbp]
    \centering
    \includegraphics[width=0.85\textwidth]{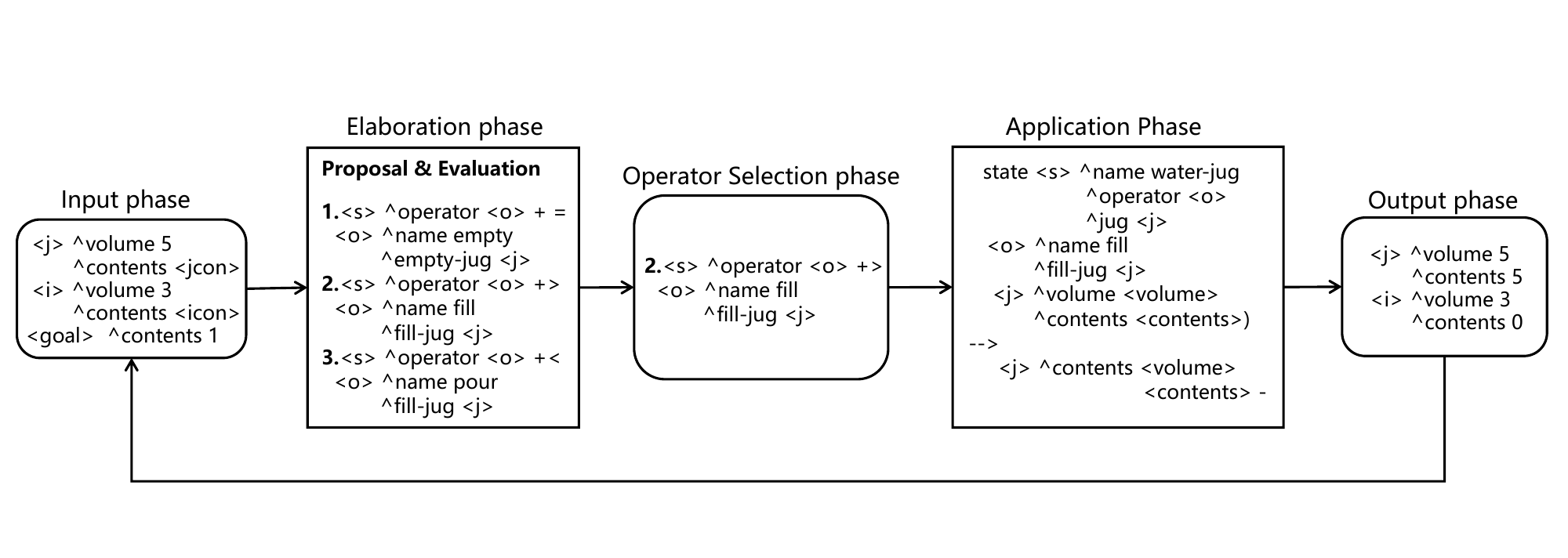}
    \caption{The decision cycle in SOAR.}
    \label{fig:soar_decision_process}
\end{figure*}

\noindent \textbf{Decision Cycle Case Study}
We use the Water Jug Problem (WJP) as a case study, with its problem space comprehensively visualized in Figure~\ref{waterjug}. Then, we elaborate on the five execution phases of the decision cycle, as detailed in Figure~\ref{fig:soar_decision_process}:
\begin{enumerate}
\item \textit{Input phase.}
This phase processes data and adds that information to the associated buffers in WM. For the WJP, this involves representing the initial state in the jugs, e.g., (0, 0) for two empty jugs, and the target goal is 1\,\text{L}.
\item \textit{Elaboration phase.}
This phase is where rules fire that elaborate the situation, propose operators, and evaluate operators. For example, within the WJP, rules would propose the \texttt{fill}, \texttt{empty}, and \texttt{pour} operators, and assign preferences of acceptable (\texttt{+=}), better (\texttt{+>}), and worse (\texttt{+<}), respectively.
\item \textit{Operator Selection phase.}
SOAR then selects one operator from the proposed candidates based on their preferences. In the WJP, if the fill (j) operator receives the highest (e.g., ``best" or uniquely ``better") preference value among all proposed operators, it would be prioritized for selection. If no operator is uniquely preferred, SOAR automatically creates a substate or subgoal to resolve the impass. Once an operator is definitively chosen, either directly or through impasse resolution, it is designated as the current operator for the next phase.
\item \textit{Application Phase.}
The selected current operator is executed. Specific apply rules for this operator fire, modifying the state in WM. For the WJP operator fill (j), an apply rule would update volume of j jug to its contents (e.g., 5\,\text{L}), transforming state (0, 0) to (5, 0).
\item \textit{Output phase.}
Once the application phase completes, the output phase sends any structures created in buffers to their respective modules. After the output phase, processing returns to the \textit{Input Phase}. For the WJP, the system then re-enters the \textit{Input Phase} with the new state (e.g., (5, 0)), ready to evaluate and select subsequent operators like pour (j, i).
\end{enumerate}

\begin{figure*}[t]
  \centering
  \includegraphics[width=0.95\textwidth]{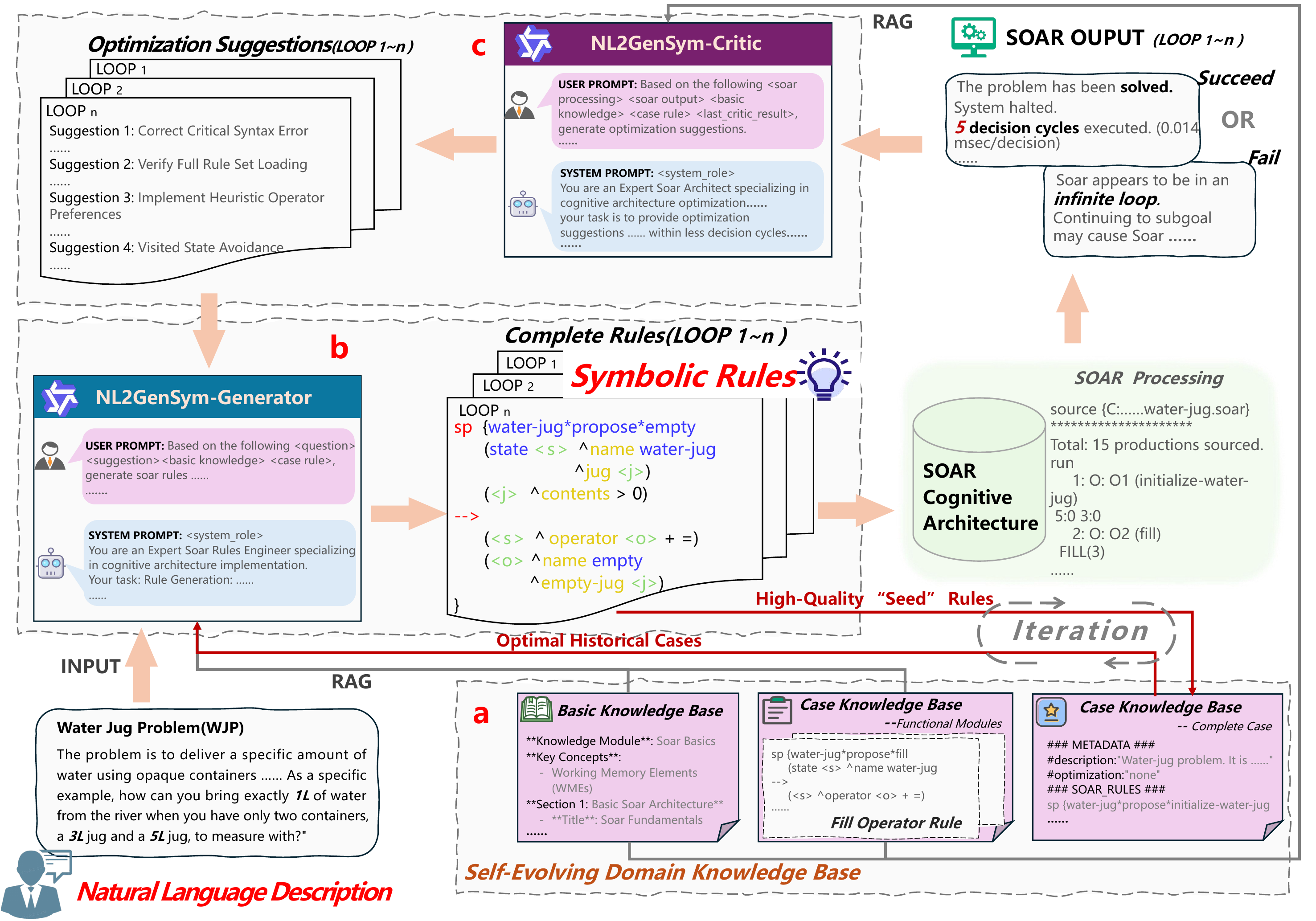}
  \caption{Operational workflow of the NL2GenSym framework, illustrated with the WJP. Natural language WJP descriptions are processed by NGS-G (b), using the Self-Evolving Domain Knowledge Base (a) via RAG, to produce symbolic rules. These rules undergo N execution cycles in the SOAR. Comprehensive feedback—including execution traces, performance data, the generated rules, and the original problem description—is then analyzed by NGS-C (c), which also leverages the Knowledge Base (a) via RAG. The NGS-C (c) outputs natural language optimization suggestions that are fed back to NGS-G (b). Additionally, the case knowledge is iteratively updated by storing currently optimal historical cases from SOAR execution as high-quality ``seed" rules within the Knowledge Base (a).}
  \label{fig:NL2GenSym framework}
\end{figure*}

\section{Method}
\label{method}

In this section, we detail the NL2GenSym framework that operates within a closed loop feedback to facilitate iterative rules generation and refinement. We begin by elaborating on the design and operational workflow of its core Execution-Grounded Generator-Critic mechanism. Following this, we present the detailed construction of the Self-Evolving Domain Knowledge Base. Finally, we delve into the specific design and implementation of the two key components: the Generator and the Critic modules.

\subsection{Overview}
\label{Overview}

To effectively integrate LLMs with SOAR for the automated generation and continuous optimization of executable symbolic rules from natural language, we have developed the NL2GenSym framework, illustrated in Figure~\ref{fig:NL2GenSym framework}.
The framework is architected around two core pillars: a Self-Evolving Domain Knowledge Base an Execution-Grounded Generator-Critic mechanism.
The knowledge base furnishes SOAR essential syntax, semantics, and a dynamically updated pool of ``seed" cases, forming a critical support structure.
The Generator-Critic mechanism then leverages this foundation. 
Its NL2GenSym-Generator~(NGS-G) is responsible for parsing high-level natural language instructions and transforming them into structured symbolic rules. These generated rules are then executed within SOAR to rigorously validate their correctness and evaluate their performance. Subsequently, the NL2GenSym-Critic~(NGS-C) analyzes these execution outcomes and the performance data, providing optimization suggestions as feedback to guide the iterative refinement of the rules by the NGS-G. These suggestions serve as direct input for subsequent rule generation cycles.

\subsection{Self-Evolving Domain Knowledge Base}
\label{KB}
Despite being trained in vast text corpora on the Internet and thus possessing broad general knowledge, LLMs often exhibit limitations in terms of depth and precision when applied to highly specialized domains such as CAs. To bridge this gap and ensure that LLMs can effectively process tasks related to the SOAR and accurately generate the corresponding rules, we have specifically designed and constructed a Self-Evolving Domain Knowledge Base~(in Figure~\ref{fig:NL2GenSym framework}a). 

The Base is structured into two main components: Basic Knowledge Base and Case Knowledge Base. The data sources cover primarily two major categories:

\begin{itemize}
\setlength{\itemsep}{0pt} 
\setlength{\parsep}{0pt} 
\setlength{\parskip}{0pt} 
\item \textit{Official and Academic Resources:} These are primarily derived from the official SOAR website\footnote{SOAR Home, \url{https://soar.eecs.umich.edu/}}, which includes its extensive tutorials, classic example programs, and published research case studies.
\item \textit{Generated Data:} These data exhibit a ``self-evolving" nature, dynamically updated and enriched with high-quality ``seed" rules generated through execution cycles.

\end{itemize}

These components systematically provide both the NGS-G and NGS-C with essential background information, detailed operational specifications, and evolving high-quality rule examples.
\subsubsection{Basic Knowledge Base}
\label{basic}

The Base is designed to provide foundational theoretical explanations and structural knowledge pertaining, intended to assist LLMs in deepening their understanding of the SOAR and to facilitate subsequent efficient knowledge retrieval. These contents, which encompass core definitions, operational principles, syntactic specifications, and the functional characteristics of key components along with their interaction mechanisms, are organized into a total of 13 distinct knowledge modules~(as detailed in Table IV, Appendix A). The construction process for this Base initially involves the systematic extraction of relevant document resources from the official website, followed by data cleaning and pre-processing. Subsequently, utilizing predefined structured templates and prompts, the processed knowledge data is classified and organized according to knowledge modules and knowledge points/elements.

\subsubsection{Case Knowledge Base}
\label{case}

The Base is designed to provide high-quality rule exemplars, aimed at helping LLMs master rule-coding paradigms and acquire generalizable rule construction techniques. The Base is divided into two sections: the Functional Module and the Complete Case. 

\textit{The Functional Module Section} focuses on providing modular rule fragments. These rules are primarily derived from some examples provided by the official website. For instance, the WJP has various strategic approaches, including basic operational rules, such as the Fill operator detailed in Table~\ref{tab:fill-rule}, policy optimization using Reinforcement Learning, Look-Ahead Search and Chunking mechanism. We deconstruct and organize these rules based on core functionalities, forming a collection of 22 distinct categories of independent, reusable functional modules. This organizational approach is intended to enable LLMs to understand and learn how to flexibly apply mechanisms of SOAR to construct solutions, thereby enhancing the richness and diversity of paths explored when generating novel rules.
\begin{table}[t]
  \centering 
  \caption{Functional Module Rule example. This rule exemplifies the fill operator functional module. It comprises two components: the propose rule and the apply rule, which together define a complete operator lifecycle. The propose rule identifies if the current state meets the preconditions for activating the fill operator. Subsequently, the apply rule defines the specific actions of operator: setting the content volume of a designated jug to its maximum capacity, thereby altering the system state.}
  \captionsetup{justification=justified,singlelinecheck=false}
  \begin{tabular}{@{}p{0.95\linewidth}@{}}
    \toprule[1.2pt]
    \multicolumn{1}{l}{\textbf{Functional Module Rule: Fill Operator}} \\
    \midrule
    \midrule
    \hspace{6em}\texttt{sp}\{water-jug*propose*fill \\
    \hspace{6em}\hspace{1em}(state $<$s$>$\;\string^name water-jug \\
    \hspace{6em}\hspace{5.6em}\string^jug $<$j$>$) \\
    \hspace{6em}\hspace{1em}($<$j$>$\;\string ^empty $>$ 0) \\
    \hspace{6em}\small$-->$ \\
    \hspace{6em}\hspace{1em}($<$s$>$\;\string ^operator$<$o$>$ +=) \\
    \hspace{6em}\hspace{1em}($<$o$>$\;\string ^name fill \\
    \hspace{6em}\hspace{3.6em}\string^fill-jug $<$j$>$)\}\\
    \\
    \arrayrulecolor{lightgray}
    \midrule
    \arrayrulecolor{black}
    \hspace{6em}\texttt{sp}\{water-jug*apply*fill\} \\
    \hspace{6em}\hspace{1em}(state $<$s$>$\;\string^name water-jug \\
    \hspace{6em}\hspace{5.6em}\string^operator $<$o$>$\\
    \hspace{6em}\hspace{5.6em}\string^jug $<$j$>$) \\
    \hspace{6em}\hspace{1em}($<$o$>$\;\string^name fill \\
    \hspace{6em}\hspace{3.6em}\string^fill-jug $<$j$>$)\\
    \hspace{6em}\hspace{1em}($<$j$>$\;\string^volume $<$volume$>$ \\
    \hspace{6em}\hspace{3.6em}\string^contents $<$contents$>$)\\
    \hspace{6em}\small$-->$ \\
    \hspace{6em}\hspace{1em}($<$j$>$\;\string^contents $<$volume$>$ \\
    \hspace{6em}\hspace{7.5em}$<$contents$>$ -)\} \\
    \bottomrule[1.2pt]
  \end{tabular}
  \label{tab:fill-rule}
\end{table}

\textit{The Complete Case Section} aims to provide executable rule exemplars, with a core focus on demonstrating the complete problem-solving process and rule organization structure from an initial state to goal achievement. 
Notably, this section is a continuously self-optimizing memory pool. During the operation of framework, the cases within this pool are iteratively updated: currently optimal historical cases identified from SOAR execution are curated as high-quality ``seed" rules and integrated back into this section of the Knowledge Base (the iteration process can be seen in Figure~\ref{fig:NL2GenSym framework}a).

This section stores a series of evolving rule cases, denoted as:

\begin{equation}
  \label{eq:case}
  C_{\text{cases}} = \{(R_i, e_i)\}_{i=0}^{N}
\end{equation}
  
- $ R_i $ denotes rules capable of completely solving the target problem, which, for $ i > 0 $, is generated by the NGS-G in the $ i $-th iteration.
  
- $ e_i $ is the performance evaluation metric obtained after executing the rules $ R_i $, such as the number of decision cycles required to complete the WJP (a lower value typically indicates superior rule performance).
    
- When $ i = 0 $, $ (R_0, e_0) $ represents the initial case. These $ R_0 $ are typically sourced from standard problem-solving solutions provided by the official source or predefined by domain experts. These rules serve as the foundational input during its initial inference (cold-start phase), aiming to substantially improve the ability of framework to generate effective and structurally valid complete rules during the initial stages, thereby mitigating the inherent uncertainty and inefficiency associated with the cold-start problem.

\subsection{NL2GenSym-Generator}
\label{NLG}

NGS-G (in Figure~\ref{fig:NL2GenSym framework}b) translates natural language problem descriptions ($ x $) into executable SOAR symbolic rules ($ R $). To optimize rule generation accuracy and efficiency, we design a comprehensive multi-source LLM prompt containing the following key components:

\begin{enumerate}
\item Task Instructions and Structured Template ($ P_{task} $): A predefined structured prompt template that incorporates specific task instructions. It serves to clearly define the generation objective while standardizing the structure of rules for compliance with SOAR syntax.
\item Retrieval-Augmented Domain Knowledge ($r_{retrieved}$): The RAG mechanism retrieves Base Knowledge Base and Functional Module content highly relevant to problem $ x $, equipping the LLM with necessary domain context.
\item Optimal Historical Cases ($ c_{\text{optiaml}} $): From the Equation~\ref{eq:case}, we select one or more complete rule sets that exhibited optimal performance in prior iterations as $ c_{\text{optiaml}} $, based on predefined performance metrics $ e_i $ (e.g., minimizing the number of decision cycles). These high-quality successful cases are injected into the prompt as few-shot cases. For the initial rule generation, the initial cases $ R_0 $ from $ C_{\text{cases}} $ are used as contextual input.
\item Optimization Suggestions from NGS-C ($S_{opt}$): Natural language feedback generated by the NGS-C to guide rule refinement. The generation process and content specifications of these suggestions are detailed in Section~\ref{sec:ngs-c}.
\end{enumerate}

Therefore, during the generation phase, the generated rules $ R $ by NGS-G can be mathematically represented as follows:

\begin{equation}
  R = \text{NGS-G} (x, P_{\text{task}}, r_{\text{retrieved}}, c_{\text{optiaml}}, S_{opt})
  \label{eq:ngs-rg-formulation}
\end{equation}

This design enables NGS-G to dynamically leverage the accumulated knowledge and historical optimal data within the framework, thereby generating more precise and task-compliant rules.

\subsection{NL2GenSym-Critic}
\label{sec:ngs-c}

NGS-C (in Figure~\ref{fig:NL2GenSym framework}c) evaluates the rules $ R $ generated by NGS-G through a multidimensional analysis framework and subsequently produces optimization suggestions \(S_{opt}\). The design leverages a customization prompt for LLMs, based on multi-dimensional information analysis, which empowers the NGS-C to conduct a comprehensive, multifaceted assessment for the generated rules. Primary inputs include:

\begin{enumerate}
\item Structured Evaluation Template (\(P_{critic}\)): Utilizing a pre-designed template, NGS-C is directed to evaluate across specific dimensions (e.g., correctness, completeness, efficiency, redundancy, readability, and adherence to SOAR best practices), generating structured outputs of evaluation results and optimization suggestions.
\item Relevant Knowledge Context ($r_{retrieved}$): Reference information retrieved via a RAG mechanism from the Basic Knowledge Base and the Functional Module, specifically for the currently generated rules \(R\).
\item Execution Trace (\(T_{exec}\)): Detailed procedural data generated when the rules \(R\) are executed within SOAR, sequence of changes in WM, and impasses generation and resolution processes. This trace provides NGS-C with a micro-level perspective to analyze dynamic behavior and potential issues within the rules.
\item Performance Metrics (\(e_i\)): Performance indicators (e.g., \(e_i\) as described in Section~\ref{case}, such as task success rate, total decision cycles, and runtime) provide a macro-level foundation for assessing the overall utility and efficiency of the rules.
\end{enumerate}

Therefore, during the evaluation phase, the optimization suggestions generated by NGS-C can be expressed in the following form:

\begin{equation}
  S_{\mathrm{opt}} = \mathrm{NGS\text{-}C} \left(x, 
    P_{\mathrm{critic}},\, R,\, r'_{\mathrm{retrieved}},\, T_{\mathrm{exec}},\, e_i 
  \right)
  \label{eq:ens-c-formulation}
\end{equation}

This design enables continuous learning through feedback, progressively enhancing rule accuracy. Ultimately, NGS-C drives closed-loop optimization within NL2GenSym by transforming evaluation insights into actionable refinements for NGS-G.

\section{Experiments and Analysis}
\label{experiments}

In this section, we present a comprehensive experimental evaluation of the NL2GenSym framework. We begin by outlining the experimental setup and key implementation details. Subsequently, we analyze the primary experimental results by evaluating the performance of NL2GenSym against baseline methods. An ablation study is then conducted to assess the impact of individual components. Furthermore, representative case studies demonstrate the ability of the framework to generate novel and efficient symbolic rules.

\subsection{Experimental Setup}

\subsubsection{Dataset}

Many complex, real-world planning tasks are conceptualized as search processes within a problem space, which is a foundational paradigm in cognitive science and artificial intelligence. The WJP serves as a canonical example of such tasks, providing a mathematically tractable abstraction of planning and reasoning challenges. As detailed in Section~\ref{sec:background}, we previously formalize problem space of the WJP and its execution through SOAR decision cycles. In our experiments, we use natural language descriptions of the WJP as inputs. The typical formulation~\cite{wray2024eliciting} follows:

\begin{myquote}
\textit{The problem is to deliver a specific amount of water using opaque containers that have no graduated markings. The amount of water that needs to be delivered will generally differ from the full capacities of the containers. As a specific example, how can you bring exactly 1\,\text{L} of water from the river when you have only two containers, a 3\,\text{L} jug and a 5\,\text{L} jug, to measure with?}
\end{myquote}

For clarity of exposition,  we formally define a WJP case as \((V_1, \dots, V_n \rightarrow V_{\text{goal}}) / Min.DC\). Here, \(V_1, \dots, V_n\) represent the contents of \(n\) jugs (with \(n=2\) for the classic WJP), all initially empty. \(V_{\text{goal}}\) denotes the target content of water to be measured in one of the jugs. \(Min.DC\) denotes the minimum decision cycles required to achieve this target state, with the initial state counted as the first decision cycle.

Evidently, an increase in \(Min.DC\) is known to cause a super-linear expansion of the problem space, significantly elevating complexity of the problem. To evaluate the generalization capabilities of our framework under varying complexities and structural differences, we also include a WJP variant involving three jugs (\(n=3\)).

Based on these considerations,  we construct a WJP dataset comprising 100 distinct cases. These cases are stratified into four distinct categories based on difficulty level, as determined \(Min.DC\): 25 easy (within 5 steps), 25 medium (between 5 and 10 steps), 30 hard (over 10 steps but not exceeding 20 steps), and 20 variant (with \(n=3\)). Table V in Appendix B contains all 100 cases.

\subsubsection{Evaluation Metrics}
\label{Metrics}

The experimental evaluation employs \textbf{Success Rate (SR)} and \textbf{Decision Cycles (DC)} metrics, inspired by recent approaches~\cite{Yujia2022Competition, Jiang2024from}. SR, the primary measure of correctness, quantifies the ability of framework to transform natural language descriptions into executable symbolic rules. Specifically, it is the percentage of the 100 cases successfully converted. DC measures efficiency by capturing the number of decision cycles. For each successfully generated rules, we execute it 100 times in SOAR and record the \textbf{Average Decision Cycles (Avg.DC)} as the quantitative efficiency metric. To provide a benchmark for evaluating Avg.DC, we define the minimum number of decision cycles required to solve each problem instance, averaged across cases, as \textbf{Average Minimum Decision Cycles (Avg.Min.DC)}.

\begin{table*}[htbp]
    \centering
    \caption{Experimental performance of different methods on our dataset. Metrics (SR, Avg.DC, Avg.Min.DC) are defined in Section~\ref{Metrics}. The best-performing result for each metric is highlighted in \textbf{boldface}. The ``$\times$" is defined as the ratio $\text{Avg.DC} / \text{Avg.Min.DC}$, quantifying rule optimization. Percentages with ``$\uparrow$'' (e.g., $\uparrow$31\%) compare performance gains against the One-Shot baseline for comparable models.}
    \label{tab:experimental_performance}
    \small
    \begin{tabularx}{0.89\textwidth}{l|cc|cccccccc}
        \toprule[1.2pt]
        \multirow{4}{*}{\textbf{Method}} &
        \multicolumn{2}{c|}{\multirow{3}{*}{\makecell[c]{\textbf{Overall} \\ \scriptsize{(Avg.Min.DC = 7.97)}}}} &
        \multicolumn{8}{c}{\textbf{Difficulty level}} \\               
        & & &
        \multicolumn{2}{c}{\makecell[c]{\textbf{easy} \\ \scriptsize{(Avg.Min.DC = 5)}}} &
        \multicolumn{2}{c}{\makecell[c]{\textbf{medium}  \\ \scriptsize{(Avg.Min.DC = 7.88)}}} &
        \multicolumn{2}{c}{\makecell[c]{\textbf{hard} \\ \scriptsize{(Avg.Min.DC = 12.86)}}} &
        \multicolumn{2}{c}{\makecell[c]{\textbf{variant} \\ \scriptsize{(Avg.Min.DC = 4.55)}}} \\ 
        &\textbf{SR$\uparrow$} &\textbf{Avg.DC$\downarrow$} &\textbf{SR} &\textbf{Avg.DC}&\textbf{SR} &\textbf{Avg.DC}&\textbf{SR} &\textbf{Avg.DC}&\textbf{SR} &\textbf{Avg.DC} \\
        \midrule
        \midrule
        Manual & 84\% & \makecell{27018.94 \\ [-0.8ex] \scriptsize{(3390.08$\times$)}} & \textbf{100\%} & \makecell{85.84 \\ [-0.8ex] \scriptsize{(17.19$\times$)}} & \textbf{100\%} & \makecell{7737.52 \\ [-0.8ex] \scriptsize{(981.92$\times$)}} & 46.67\% & \makecell{147984.64 \\ [-0.8ex] \scriptsize{(11507.36$\times$)}}& \textbf{100\%} & \makecell{111.1 \\ [-0.8ex] \scriptsize{(24.42$\times$)}}  \\
        \midrule
        \multicolumn{11}{l}{\bfseries\itshape gemini-2.5-pro-exp-03-25}\\[5pt]
        Zero-Shot& 0 & None & 0 & None & 0 & None & 0 & None & 0 & None \\[5pt]
        \arrayrulecolor{lightgray}
        \midrule
        \arrayrulecolor{black}
        One-Shot& 55\% & \makecell{20344.47 \\[-0.8ex] \scriptsize{(2552.63$\times$)}} & 76\% & \makecell{169.21 \\ [-0.8ex] \scriptsize{(33.84$\times$)}} & 76\% & \makecell{8428.89 \\[-0.8ex]  \scriptsize{(1069.66$\times$)}} & 20\% & \makecell{118192.83 \\ [-0.8ex] \scriptsize{(9190.73$\times$)}}& 55\% & \makecell{215.64 \\ [-0.8ex] \scriptsize{(47.39$\times$)}}  \\[5pt]
        \arrayrulecolor{lightgray}
        \midrule
        \arrayrulecolor{black}
        \bfseries
        \makecell[l]{NL2GenSym \\ [-0.8ex] \tiny{(\itshape Gemini-2.5(FP))}}&
        \bfseries \makecell{86\% \\ [-0.8ex] \scriptsize{($\uparrow$31\%)}}  &
        \bfseries \makecell{14.06 \\ [-0.8ex] \scriptsize{(1.76$\times$)}} &
        \bfseries \makecell{\textnormal{96\%} \\ [-0.8ex] \scriptsize{($\uparrow$20\%)}}&
        \bfseries \makecell{11.32 \\ [-0.8ex] \scriptsize{(2.26$\times$)}} &
        \bfseries \makecell{\textnormal{92\%} \\ [-0.8ex] \scriptsize{($\uparrow$16\%)}} &
        \bfseries \makecell{14.80 \\ [-0.8ex] \scriptsize{(1.88$\times$)}} &
        \bfseries \makecell{76.67\% \\ [-0.8ex] \scriptsize{($\uparrow$56.67\%)}} &
        \bfseries \makecell{18.29 \\ [-0.8ex] \scriptsize{(1.42$\times$)}}&
        \bfseries \makecell{\textnormal{80\%} \\ [-0.8ex] \scriptsize{($\uparrow$25\%)}} &
        \bfseries \makecell{11.01 \\ [-0.8ex] \scriptsize{(2.42$\times$)}}  \\
        \midrule
        \multicolumn{11}{l}{\bfseries\itshape qwen-max-latest}\\[5pt]
        Zero-Shot& 0 & None & 0 & None & 0 & None & 0 & None & 0 & None \\[5pt]
        \arrayrulecolor{lightgray}
        \midrule
        \arrayrulecolor{black}
        One-Shot& 66\% & \makecell{20669.56 \\ [-0.8ex] \scriptsize{(2593.42$\times$)}} & 84\% & \makecell{222.71 \\ [-0.8ex] \scriptsize{(44.54$\times$)}} & 72\% & \makecell{19178.67 \\ [-0.8ex] \scriptsize{(2433.84$\times$)}} & 46.67\% & \makecell{134099.14 \\ [-0.8ex] \scriptsize{(10427.62$\times$)}}& 85\% & \makecell{156.88 \\ [-0.8ex] \scriptsize{(34.48$\times$)}}  \\[5pt]
        \arrayrulecolor{lightgray}
        \midrule
        \arrayrulecolor{black}
        \bfseries 
        NL2GenSym &  
        \bfseries \makecell{91\% \\ [-0.8ex] \scriptsize{($\uparrow$25\%)}} &
        \bfseries \makecell{15.81 \\ [-0.8ex] \scriptsize{(1.98$\times$)}} &
        \bfseries \makecell{\textnormal{96\%} \\ [-0.8ex] \scriptsize{($\uparrow$12\%)}} &
        \bfseries \makecell{13.33 \\ [-0.8ex] \scriptsize{(2.67$\times$)}} &
        \bfseries \makecell{\textnormal{92\%} \\ [-0.8ex] \scriptsize{($\uparrow$20\%)}} &
        \bfseries \makecell{12.91 \\ [-0.8ex] \scriptsize{(1.64$\times$)}} &
        \bfseries \makecell{86.67\% \\ [-0.8ex] \scriptsize{($\uparrow$40\%)}} &
        \bfseries \makecell{16.69 \\ [-0.8ex] \scriptsize{(1.3$\times$)}} &
        \bfseries \makecell{\textnormal{90\%} \\ [-0.8ex] \scriptsize{($\uparrow$5\%)}} &
        \bfseries \makecell{21.56 \\ [-0.8ex] \scriptsize{(4.74$\times$)}} \\
        \bottomrule[1.2pt]
    \end{tabularx}
\end{table*}

\subsubsection{Baseline Methods}
\label{Baseline}

To comprehensively evaluate the effectiveness of our framework, we design three baseline experiment categories for comparative analysis. All symbolic rules generated by these baselines and NL2GenSym are executed uniformly within SOAR, with execution results meticulously recorded.

\begin{enumerate}
\item \textit{Manual Encoding.} Establishes a fundamental benchmark using simple and basic symbolic rules crafted manually. This represents traditional rule-based problem solving.
\item \textit{LLM Zero-Shot.} Evaluates the inherent capability of LLMs to translate natural language task prompts directly into executable symbolic rules without exemplars. Assesses novel rule generation potential without prior examples.
\item \textit{LLM One-Shot.} Enhances zero-shot generation by incorporating a single exemplar (identical to Manual Encoding rules) with the task prompt. Measures the ability of LLMs to improve rule quality and efficiency through contextual guidance.
\end{enumerate}

\subsection{Implementation details}

All experiments in this paper are conducted on a laptop. It is equipped with an i9-13980HX CPU and an GeForce RTX 4060, running Windows 11. To reinforce the practical implementation of our work, we elucidate the following aspects:

\begin{enumerate}
\item \textit{Manual Rules.}
The manual encoding rules employed in the Manual baseline represent a fundamental, non-optimized rule set. Crucially, this same rule set serves dual purposes: as the exemplar case for the LLM-OneShot experiment, and as the initial case for our proposed framework's experiments.
\item \textit{Execution Termination and Failure Adjudication.}
Experiments reveal that rules with high minimum cycle requirements (Avg.Min.DC \texttt{>} 10) exhibit heightened susceptibility to infinite execution loops.
These cases compromise both practical runtime feasibility and comparative analysis validity. We therefore implement a strict termination protocol: any execution exceeding 500,000 decision cycles without achieving the target state (e.g., 1L in 5L jug) is terminated and record as a failure.
\item \textit{RAG Implementation Details.}  
Our RAG implementation uses DPR dual encoders~\cite{karpukhin2020dense} with FAISS indexing~\cite{johnson2019billion} for efficient retrieval from the Self-Evolving Domain Knowledge Base. For each query (NGS-G/NGS-C), we retrieve \textit{top-5} chunks from the Basic Knowledge Base and \textit{top-10} chunks from the Functional Module Section, augmenting the LLM prompts accordingly.
\item \textit{LLMs Configuration.}
Model selection prioritized state-of-the-art capabilities based on Hugging Face's Chatbot Arena Leaderboard. We select top-ranked representative Gemini and Qwen series models. Furthermore, to comprehensively evaluate the superiority of the method proposed herein, we also incorporated a comparatively smaller-parameter LLM of the same type to serve as a comparative baseline, thereby facilitating a more nuanced assessment of efficacy about our method. Specific versions and temperature configurations are provided in Table VI, Appendix B.
\end{enumerate}

\subsection{Comparisons with Baselines}
\label{result_exp}

To assess the efficacy of the proposed NL2GenSym framework, we conduct a comparative analysis against three established baselines. This evaluation primarily focuses on two key performance metrics detailed in Section~\ref{Metrics}: \textbf{\textit{Success Rate (SR)}} and \textbf{\textit{Average Decision Cycles (Avg.DC)}},  with comprehensive results summarized in Table~\ref{tab:experimental_performance}. Building upon these findings, we subsequently analyze the mechanisms driving the observed phenomena.

\subsubsection{Manual Baseline Performance Analysis}

The Manual Encoding method achieves success in easy, medium, and variant cases but exhibits markedly low execution efficiency. Its Avg.DC consistently far exceeds the Avg.Min.DC: on the overall dataset, the former is typically over 1000$\times$ greater than the latter; even for easy cases, Avg.DC often exceeds Avg.Min.DC by more than 10$\times$. This efficiency bottleneck becomes critical in hard cases, where the method frequently fails due to exceeding the computational limit (500,000 decision cycles, as detailed in Section~\ref{Baseline}), resulting in an SR of only 46.67\%.

Experimental data for the Manual Encoding baseline reveal a sharp increase in decision cycles with rising case complexity. This directly reflects core problem-solving mechanism of SOAR: abstraction as search within a problem space. 
Consequently, as the minimum required decision cycles (i.e., the optimal steps) increase, the problem space expands super-lineally. This exponential growth causes traditional manual encoding methods to fail in converging to solutions within acceptable timeframes for complex problems. Performance consequently becomes critically dependent on: (1) developers' depth of domain-specific knowledge, and (2) their understanding of problem-space intricacies. Performance improvements require iterative manual optimization, which constitutes a labor-intensive process with poor scalability. \textbf{\itshape These limitations expose fundamental scalability and adaptability constraints while highlighting prohibitive knowledge engineering barriers and human capital requirements.}

\subsubsection{LLMs~(Zero-Shot \& One-Shot) Baseline Performance Analysis}

In zero-shot experiments, the employed LLMs consistently fail to translate natural language descriptions into executable symbolic rules. The introduction of an exemplar in one-shot experiments yield discernible improvements in task comprehension, elevating SR to 55\%--66\%. However, this performance gain prove insufficient for hard cases, where SR declines to as low as 20\%. More critically, successful one-shot executions fail to improve efficiency, with Avg.DC maintaining parity with manual baseline performance.

Experimental data from LLM baselines validate their proficiency in natural language understanding yet reveal significant limitations when applied to domains requiring specialized structured knowledge. \textbf{\itshape This challenge originates in their core design: as sequence transducers and probabilistic predictors, LLMs rely on statistical correlations rather than formal logic~\cite{2023Generating}}. Consequently, when generating outputs demanding strict syntax and operational logic (e.g., SOAR rules), these models frequently produce non-compliant or ineffectively executable results~\cite{2024Exploring}.

\subsubsection{NL2GenSym Framework Performance Analysis}

NL2GenSym demonstrates robust capabilities in both generating executable rules from natural language and optimizing their efficiency. The framework achieves 86\%+ SR across the WJP dataset, indicating strong capacity for producing syntactically correct and structurally complete rules. Compared to the One-Shot baseline, NL2GenSym shows \textgreater25\% SR improvement overall, with particularly notable gains (56.67\% increase) on hard cases using Gemini models.
In rule optimization, NL2GenSym significantly outperforms all baselines. While manual and LLM methods exhibit Avg.DC orders of magnitude above optimal levels, our framework consistently constrains Avg.DC within 2$\times$ Avg.Min.DC. In hard cases, efficiency improves further as the framework achieves an Avg.DC consistently lower than 1.5$\times$ Avg.Min.DC.

The performance superiority of NL2GenSym originates in its architectural design, which overcomes limitations of direct LLMs generation and manual encoding. Significant SR improvements derive from: (1) RAG-enabled Self-Evolving Domain Knowledge Base that supplies NGS-G with critical SOAR context/exemplars, reducing rule-generation flaws;  
(2) The NGS-C, through its closed-loop interaction with SOAR execution, validates rules and provides targeted natural language feedback for iterative correction and integrity enhancement.

The iterative Execution-Grounded Generator-Critic optimization cycle is also responsible for the remarkable reduction in Avg.DC. \textbf{\itshape 
NL2GenSym facilitates an emergent learning process. In this process, the NGS-C analyzes execution-grounded feedback from the SOAR environment, guiding the NGS-G to discover and internalize novel, high-efficiency heuristic rules.} The continuous, data-driven strategy optimization, largely absent in baseline methods, explains their comparatively poor efficiency. A detailed analysis of such generative symbolic rules for two representative cases is presented in Section~\ref{sub:generative}, further substantiating this emergent optimization capability.

\subsubsection{The Potential of NL2GenSym to Mitigate Model Scale Dependencies}

A key finding stems from an experiment highlighting the power of structured guidance: We compare our NL2GenSym framework using the smaller-parameter model (gemini-2.5-flash-preview-04-17) to the standard one-shot prompting baseline using the significantly larger Gemini-Pro model (gemini-2.5-pro-exp-03-25). As Table~\ref{tab:experimental_performance} demonstrates, a performance inversion occurs. Despite its smaller scale, the Gemini-Flash model guided by our framework achieves 86\% SR and near-optimal reasoning efficiency (1.76$\times$ Avg.Min.DC). This significantly surpasses the results from the larger Gemini-Pro using one-shot prompting, which achieves only 55\% SR with an efficiency of 2500$\times$ Avg.Min.DC on the overall dataset.

This outcome reveals that the performance leap is not solely attributable to model scale but significantly stems from the structured guidance of the NL2GenSym architecture. The Execution-Grounded Generator-Critic optimization cycle effectively overcomes the inherent limitations of smaller models.
\textbf{\itshape Crucially, this finding (while preliminary) suggests that for problems requiring verifiable, structured symbolic outputs, a well-designed neuro-symbolic framework can be more impactful than sheer model scale.} It therefore highlights a promising path towards developing high-performance, more efficient, and accessible AI systems, offering an attractive alternative to the widespread reliance on resource-intensive ultra-large-scale models.

\subsection{Ablation Study}
\label{ablation_study}

\begin{figure}[htbp]
  \centering
  \includegraphics[width=\linewidth]{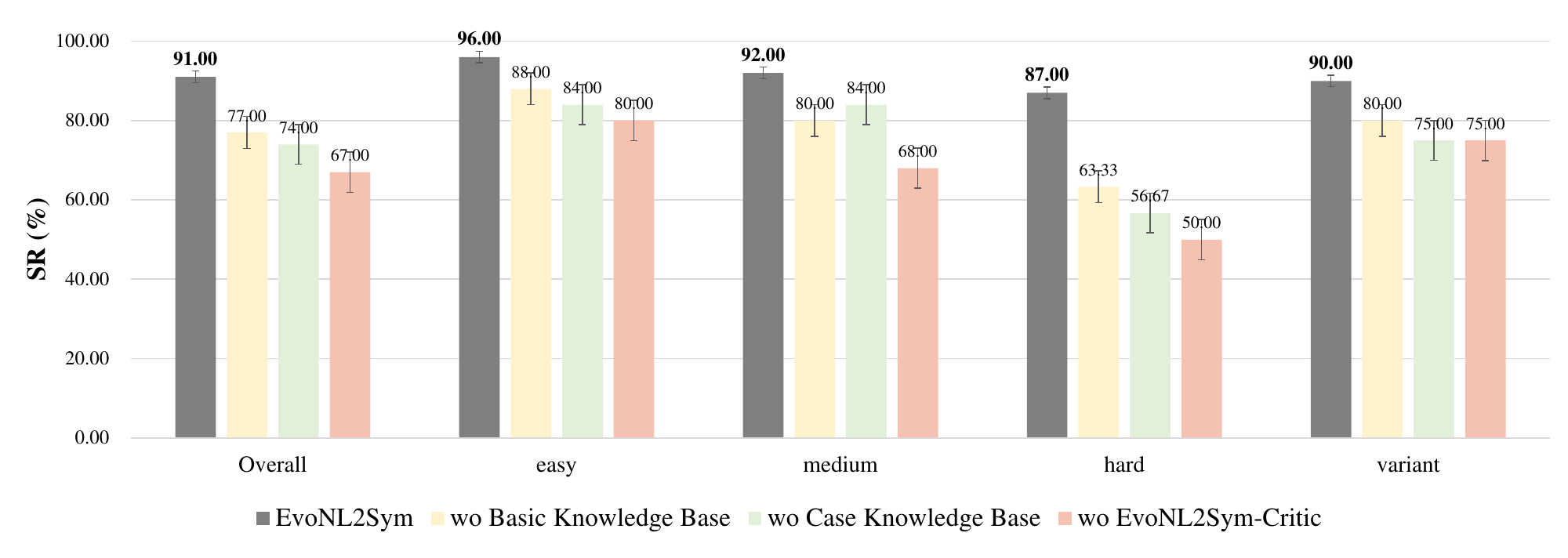}
  \caption{Results of ablation studies for Success Rate (\%) metrics.}
  \label{fig:Ablation-SR}
\end{figure}

To evaluate the contribution of core components in our NL2GenSym framework, we conduct ablation studies on three key elements: (1) the Basic Knowledge Base, (2) the Case Knowledge Base (with initial case intentionally retained to ensure a foundational baseline, as LLMs perform poorly in zero-shot settings—see Section~\ref{result_exp}), and (3) the NGS-C.
Experimental configurations are detailed in Table VI, Appendix B, with ablation results present in Figure~\ref{fig:Ablation-SR} and Table~\ref{tab:Ablation-ds}.

We find that the ablation of the NGS-C exerted the most substantial impact on performance of the framework.
The removal of the Critic yields the most pronounced decrease in SR, with a 23\% reduction observed across the overall dataset. 
Concurrently, its Avg.DC increases markedly, exceeding 1,000$\times$ Avg.Min.DC. Furthermore, this SR degradation proves especially acute for challenging and variable cases, where the SR drops by 37\%.
This result underscores the pivotal importance of the feedback mechanism driven by the NGS-C. This mechanism comprehensively evaluates the entire rule generation lifecycle—from initial inputs and generated rules to SOAR execution traces and performance metrics—while providing targeted refinement guidance. This end-to-end evaluation proves fundamental for optimizing final solution quality. Without this effective feedback loop, the NGS-G consistently struggles to autonomously produce novel, high-quality symbolic rules.

Ablating the Basic Knowledge Base yields a less severe SR impact than removing the NGS-C, yet still causes a notable performance decline—manifested by a 17\% SR reduction. In decision efficiency metrics, the absence of the Basic Knowledge Base does not reduce Avg.DC, maintaining values around 1,000$\times$ Avg.Min.DC. Critically, however, this foundational domain knowledge gap hinders the learning of effective optimization strategies. These results collectively underscore the role of Basic Knowledge Base as the essential cornerstone for symbolic rule construction.

Similarly, ablating the Case Knowledge Base yields a comparable SR reduction of 14\%, even when retaining the initial case. Moreover, the absence of experiential guidance from the Case Knowledge Base diminishes solution space exploration efficiency, with Avg.DC remaining unoptimized at approximately 1,000$\times$ Avg.Min.DC. These results collectively confirm the efficacy of case-based learning and experience transfer in accelerating and refining symbolic rule discovery.

In summary, these ablation studies affirm that the Basic Knowledge Base, Case Knowledge Base and the NGS-C operate not merely as additive components but as synergistic elements. The Basic Knowledge Base provides essential SOAR-specific grounding that improves rule correctness; the Case Knowledge Base offers crucial experiential guidance to accelerate effective pattern discovery; and the NGS-C drives iterative refinement necessary for both error correction and high-efficiency heuristic generation. Together, these components form an indispensable triad that underpins efficacy in rule generation and achieved efficiency in problem-solving within the NL2GenSym framework.

\begin{table}[htbp]
  \centering
  \footnotesize
  \setlength{\tabcolsep}{5pt} 
  \renewcommand{\arraystretch}{1.2} 
  \caption{Results of ablation studies for the ratio ($\times$) of Avg.DC to Avg.Min.DC metrics.}
  \label{tab:Ablation-ds}
  \begin{tabular}{
    l|
    S[table-format=4.2]|
    S[table-format=2.2]
    S[table-format=2.2]
    S[table-format=4.2]
    S[table-format=2.2]
  }
    \toprule[1.1pt]
    \textbf{Method}                  & \textbf{Overall} & \textbf{easy} & \textbf{medium} & \textbf{hard}    & \textbf{variant} \\
    \midrule
    \midrule
    NL2GenSym               & 1.98      & 2.67   & 1.64     & 1.30      & 4.74      \\
    \midrule
    w/o Basic KB & 3711.45   & 77.15  & 86.63    & 9224.20   & 23.32     \\
        \arrayrulecolor{lightgray}
        \midrule
        \arrayrulecolor{black}
    w/o Case KB  & 2800.71   & 57.34  & 71.95    & 7463.39   & 32.63     \\
        \arrayrulecolor{lightgray}
        \midrule
        \arrayrulecolor{black}
    w/o NGS-C     & 2948.71   & 53.86  & 81.73    & 8070.45   & 21.36     \\
    \bottomrule[1.1pt]
  \end{tabular}
\end{table}

\subsection{Representative Case Studies}
\label{sub:generative}

Our experimental findings demonstrate that NL2GenSym framework can effectively generate and optimize problem-solving strategies through its iterative process. This leads to the emergence of novel optimization methods that significantly enhance solution efficiency. To illustrate this capability, we present an analysis of two representative cases. The detailed optimization suggestions formulated by NGS-C and the corresponding generated rules are provided in Appendix C.

\vspace{1em}
\noindent \textbf{Case 1:} Using the Gemini model for problem \textit{(7,9$\rightarrow$1)/13 (Case 70)}, the strategy generated by the framework in the \textit{17th iteration} optimizes the average decision cycles to \textit{17.5 (1.35$\times$)}. This strategy integrates rules for calculating the distance between the current state and the goal state, as well as rules for predicting the resultant states after operator execution (e.g., fill, empty, pour). Crucially, it leverages these predictions to prohibit transitions to previously visited state configurations, thereby effectively pruning the search space. Concurrently, it employs a goal-oriented heuristic to prioritize operators capable of achieving the target state in a single step.

\vspace{1em}
\noindent \textbf{Case 2:} Using the Qwen model for problem \textit{(4,9$\rightarrow$2)/11 (Case 53)}, the strategy generated in the \textit{9th iteration} optimizes the average decision cycles to \textit{13 (1.18$\times$)}. This strategy employs a heuristic-based operator proposal and preference selection mechanism, assigning differential preference levels to various operations: e.g., prioritizing filling the smaller capacity jug, assigning neutral preference to ``pour" operations, and discouraging the emptying of the larger capacity jug. This well-designed preference hierarchy guides the decision-making mechanism towards more promising solution pathways, thereby reducing ineffective exploration of suboptimal branches. Furthermore, this strategy further prunes the search space by explicitly rejecting invalid operations (e.g., attempting to empty an already empty jug), eliminating irrational action sequences.

\vspace{1em}
These cases concretely illustrate the capacity of NL2GenSym to autonomously discover diverse and sophisticated heuristics, from direct search-space pruning (e.g., ``visited-state avoidance") to strategic preference setting. This ability is particularly promising for real-world engineering applications, as these emergent strategies mirror fundamental efficiency principles such as preventing redundant operations in logistics or design processes.

\section{Discussion}
\label{discussion}

While NL2GenSym demonstrates significant efficacy, we acknowledge several limitations that warrant discussion. 
Firstly, our validation on the WJP, a well-defined theoretical task, may not fully capture the complexities of real-world applications, which often involve ambiguous problem statements and noisy, dynamic contexts. Bridging this theory-to-reality gap is a critical next step. 
Secondly, the iterative nature of our Execution-Grounded Generator-Critic loop, while powerful, incurs substantial computational costs and does not always yield a linear or monotonic improvement in rule quality. This highlights a core challenge in optimizing the efficiency and predictability of the refinement process. 
Finally, while NL2GenSym enables LLMs to generate functional SOAR rules, the depth of their genuine comprehension of the underlying principles of SOAR remains an open question. Although our case studies demonstrate the potential of the framework to produce optimized and interpretable heuristics, a more comprehensive analysis is required to fully substantiate this capability across a broader range of problems.

\section{Related Work} 
\label{work}

\noindent \textbf{LLMs for Cognitive Architectures}
The advent of LLMs~\cite{achiam2023gpt, brown2020language, vaswani2017attention} has revitalized CAs by offering novel solutions for knowledge acquisition and human-like interaction~\cite{sumers2023cognitive}. Researchers are actively integrating LLMs with CAs like SOAR and ACT-R in several key ways. One approach employs LLMs as ``front-ends" or ``knowledge augmentation modules": Laird et al.~\cite{laird2023proposal} further proposed integrating Transformers, trained online using an agent's experiences, as a novel learning and memory component within a CA like SOAR, aiming to learn context-sensitive, similarity-based predictive knowledge from internal agent experiences rather than just external language. Kirk et al.~\cite{kirk2023exploiting} investigated LLMs as external knowledge sources, with their STARS strategy analyzing, repairing, and selecting LLM responses to acquire knowledge consistent with an agent's context. Wray et al.~\cite{wray2024eliciting} explored using LLMs to translate natural language problem descriptions into semi-formal problem space specifications, thereby assisting CAs in establishing initial task models. Another significant direction utilizes LLMs for higher-level reasoning and planning within CA-driven systems. Yao et al.~\cite{yao2023react} proposed ReAct framework for synergizing reasoning and acting in large language models. Liu et al.~\cite{liu2025large} developed GM-Agent, where LLMs process textualized world models and utilize generative memory for task planning. Sharma et al.~\cite{sharma2025novel} presented a novel Voice in Head (ViH) framework, that integrated LLMs and the power of semantic understanding to enhance robotic navigation and interaction within complex environments. Zargarzadeh et al.~\cite{zargarzadeh2025decision} proposed a multi-modal LLM integration in robot-assisted surgery for autonomous blood suction. These research showcased the potential of LLMs as abstract ``thinkers" and ``planners". Furthermore, LLMs are being explored to assist in CA model development itself, Wu et al.~\cite{wu2023comparing, wu2024cognitive} demonstrated their use as conversational interfaces for ACT-R/SOAR model creation, highlighting the necessity of iterative prompting and debugging due to LLM challenges with CA-specific syntax/semantics, and proposed prompt patterns for improved interaction.

{\itshape While prior works primarily use LLMs to assist CAs, our framework integrates the LLM at a deeper level, empowering it to generate the core executable logic of the CA. Crucially, this integration is not one-way; it forms a closed-loop where execution feedback from the CA drives the iterative refinement of the generative process in the LLMs. This approach moves beyond mere augmentation, enabling emergent discovery of optimized cognitive strategies.}

\noindent \textbf{LLMs for Symbolic Knowledge Generation}
The remarkable generative capabilities of LLMs have spurred significant interest in their application to symbolic knowledge generation, extending beyond natural language text to more structured domains such as semantic parsing~\cite{jia2016data, andreas2019good}. While conventional approaches often relied on fine-tuned generative models paired with grammar-based parsers for symbolic outputs~\cite{zhong2020grounded, guo2021revisiting}, recent efforts increasingly leverage LLMs directly. For instance, Wang et al.~\cite{wang2021codet5} presented CodeT5, an identifier-aware unified model that excels at generating code by understanding its semantics. Another line of work employs LLMs to generate synthetic data containing symbolic knowledge to train other models. Considering the difficulty in designing grammar to sample useful symbolic forms in complex domains, Yang et al.~\cite{yang2022addressing} assumed access to an unlabeled corpus of symbolic language, which was represented in canonical forms, and simulates natural language inputs via LLMs. Rosenbaum et al.~\cite{rosenbaum2022clasp}, with their CLASP method, used an LLM to create data for few-shot cross-lingual semantic parsing, thereby enhancing smaller model performance. These advancements highlighted potential of LLMs in automating the creation of diverse symbolic knowledge, though challenges in ensuring correctness, consistency, and adherence to complex constraints remain critical areas of research.

{\itshape A central challenge in this domain is ensuring the functional correctness and constraint adherence of the generated symbols. Our framework directly confronts this by using the SOAR execution environment itself as a dynamic, post-generation validator. This execution-grounded feedback ensures the generated rules are not only syntactically valid but functionally effective, addressing a key limitation of open-loop or grammar-based generation methods.}

\section{Conclusion and Future Work}
\label{conclusion}
In this paper, we explore the critical challenges of integrating LLMs with SOAR, particularly in achieving a practical, end-to-end solution from natural language to executable symbolic rules. We propose NL2GenSym, a novel framework that, to our knowledge, represents the first successful end-to-end LLM-SOAR integration. 
It offers a new paradigm for automated problem-solving by enabling direct generation of symbolic rules from natural language, thereby lowering the barrier to SOAR adoption, especially in knowledge-scarce scenarios. 
Experimental validation on the WJP confirms the generative capacity of the framework. It not only achieves a rule generation success rate exceeding 86\%, but also facilitates the emergent discovery of novel, high-efficiency heuristics that reduce decision cycles to near-optimal levels (1.98× theoretical optimum). Furthermore, we show that the framework can empower smaller-parameter models to surpass their larger counterparts, highlighting that a well-designed architecture is more critical than sheer model scale. Our work thus offers a viable methodology for autonomously transforming natural language into generative symbolic knowledge, advancing the capabilities of intelligent agents.

There are several promising directions for future research based on our work: (1) The WJP serves as a theoretical benchmark, subsequent research requires validation of the framework on practical datasets from domains like logistics planning or robotic task execution to assess real-world utility and robustness; (2) Future work focuses on enhancing iterative learning efficiency through prompt engineering optimization, refinement of domain knowledge bases, and improved RAG invocation mechanisms. These enhancements shorten the LLMs iterative learning cycle to address complex tasks with stringent timeliness demands; (3) Deepening research into LLM adaptability involves investigating model-adaptive prompting strategies and domain knowledge integration paradigms. These investigations target the varying sensitivities and learning efficacies across LLMs, aiming to enhance universality and optimization effectiveness.

\bibliographystyle{IEEEtran}
\bibliography{main.bib}

\begin{thebibliography}{10}
\providecommand{\url}[1]{#1}
\csname url@samestyle\endcsname
\providecommand{\newblock}{\relax}
\providecommand{\bibinfo}[2]{#2}
\providecommand{\BIBentrySTDinterwordspacing}{\spaceskip=0pt\relax}
\providecommand{\BIBentryALTinterwordstretchfactor}{4}
\providecommand{\BIBentryALTinterwordspacing}{\spaceskip=\fontdimen2\font plus
\BIBentryALTinterwordstretchfactor\fontdimen3\font minus \fontdimen4\font\relax}
\providecommand{\BIBforeignlanguage}[2]{{%
\expandafter\ifx\csname l@#1\endcsname\relax
\typeout{** WARNING: IEEEtran.bst: No hyphenation pattern has been}%
\typeout{** loaded for the language `#1'. Using the pattern for}%
\typeout{** the default language instead.}%
\else
\language=\csname l@#1\endcsname
\fi
#2}}
\providecommand{\BIBdecl}{\relax}
\BIBdecl

\bibitem{newell1994unified}
A.~Newell, \emph{Unified theories of cognition}.\hskip 1em plus 0.5em minus 0.4em\relax Harvard University Press, 1994.

\bibitem{laird2012soar}
J.~E. Laird, \emph{The Soar Cognitive Architecture}.\hskip 1em plus 0.5em minus 0.4em\relax MIT Press, 2012.

\bibitem{wooldridge1994agent}
M.~Wooldridge and N.~R. Jennings, ``Agent theories, architectures, and languages: a survey,'' in \emph{International Workshop on Agent Theories, Architectures, and Languages}.\hskip 1em plus 0.5em minus 0.4em\relax Springer, 1994, pp. 1--39.

\bibitem{rosenbloom1985r1}
P.~S. Rosenbloom, J.~E. Laird, J.~McDermott, A.~Newell, and E.~Orciuch, ``R1-soar: An experiment in knowledge-intensive programming in a problem-solving architecture,'' \emph{IEEE Transactions on Pattern Analysis and Machine Intelligence}, no.~5, pp. 561--569, 1985.

\bibitem{kotseruba202040}
I.~Kotseruba and J.~K. Tsotsos, ``40 years of cognitive architectures: core cognitive abilities and practical applications,'' \emph{Artificial Intelligence Review}, vol.~53, no.~1, pp. 17--94, 2020.

\bibitem{laird1986chunking}
J.~E. Laird, P.~S. Rosenbloom, and A.~Newell, ``Chunking in soar: The anatomy of a general learning mechanism,'' \emph{Machine learning}, vol.~1, pp. 11--46, 1986.

\bibitem{muni2023better}
M.~K. Muni, S.~Kumar, C.~Sahu, P.~R. Dhal, D.~R. Parhi, and S.~K. Patra, ``Better decision-making strategy with target seeking approach of humanoids using hybridized soarann-fuzzy technique,'' \emph{Journal of Computational Science}, vol.~70, p. 102026, 2023.

\bibitem{ramzani2024recognition}
M.~Ramzani~Shahrestani, S.~Motamed, and M.~Yamaghani, ``Recognition of facial emotion based on soar model,'' \emph{Frontiers in Neuroscience}, vol.~18, p. 1374112, 2024.

\bibitem{horawalavithana2023scitune}
S.~Horawalavithana, S.~Munikoti, I.~Stewart, and H.~Kvinge, ``Scitune: Aligning large language models with scientific multimodal instructions,'' \emph{arXiv preprint arXiv:2307.01139}, 2023.

\bibitem{hagos2024recent}
D.~H. Hagos, R.~Battle, and D.~B. Rawat, ``Recent advances in generative ai and large language models: Current status, challenges, and perspectives,'' \emph{IEEE Transactions on Artificial Intelligence}, 2024.

\bibitem{adornetto2025generative}
C.~Adornetto, A.~Mora, K.~Hu, L.~I. Garcia, P.~Atchade-Adelomou, G.~Greco, L.~A.~A. Pastor, and K.~Larson, ``Generative agents in agent-based modeling: Overview, validation, and emerging challenges,'' \emph{IEEE Transactions on Artificial Intelligence}, 2025.

\bibitem{valmeekam2023planbench}
K.~Valmeekam, M.~Marquez, A.~Olmo, S.~Sreedharan, and S.~Kambhampati, ``Planbench: An extensible benchmark for evaluating large language models on planning and reasoning about change,'' \emph{Advances in Neural Information Processing Systems}, vol.~36, pp. 38\,975--38\,987, 2023.

\bibitem{xie2024travelplanner}
J.~Xie, K.~Zhang, J.~Chen, T.~Zhu, R.~Lou, Y.~Tian, Y.~Xiao, and Y.~Su, ``Travelplanner: A benchmark for real-world planning with language agents,'' \emph{arXiv preprint arXiv:2402.01622}, 2024.

\bibitem{pallagani2024prospects}
V.~Pallagani, B.~C. Muppasani, K.~Roy, F.~Fabiano, A.~Loreggia, K.~Murugesan, B.~Srivastava, F.~Rossi, L.~Horesh, and A.~Sheth, ``On the prospects of incorporating large language models (llms) in automated planning and scheduling (aps),'' in \emph{Proceedings of the International Conference on Automated Planning and Scheduling}, vol.~34, 2024, pp. 432--444.

\bibitem{laird2023proposal}
J.~E. Laird, R.~E. Wray, S.~Jones, J.~R. Kirk, and P.~Lindes, ``Proposal for cognitive architecture and transformer integration: online learning from agent experience,'' in \emph{Proceedings of the AAAI Symposium Series}, vol.~2, no.~1, 2023, pp. 302--306.

\bibitem{kirk2023exploiting}
J.~R. Kirk, R.~E. Wray, and J.~E. Laird, ``Exploiting language models as a source of knowledge for cognitive agents,'' in \emph{Proceedings of the AAAI Symposium Series}, vol.~2, no.~1, 2023, pp. 286--294.

\bibitem{wu2023comparing}
S.~Wu, R.~F. Souza, F.~E. Ritter, and W.~T. Lima~Jr, ``Comparing llms for prompt-enhanced act-r and soar model development: A case study in cognitive simulation,'' in \emph{Proceedings of the AAAI Symposium Series}, vol.~2, no.~1, 2023, pp. 422--427.

\bibitem{lewis2020retrieval}
P.~Lewis, B.~Doll{\'a}r{\'y}, M.~Goldstein, A.~Joshi, Y.~Dou, M.~Iyyer, R.~Paulus, R.~Nallapati, and A.~McCallum, ``Retrieval-augmented generation for knowledge-intensive nlp tasks,'' \emph{Advances in Neural Information Processing Systems}, vol.~33, pp. 9459--9474, 2020.

\bibitem{Supparesk2025Chatbot}
S.~Rittikulsittichai and T.~Siriborvornratanakul, ``An intelligent chatbot assistant for comprehensive troubleshooting guidelines and knowledge repository in printed circuit board production,'' \emph{IEEE Transactions on Artificial Intelligence}, vol.~6, 2025.

\bibitem{newell1993formulating}
A.~Newell, G.~R. Yost, J.~E. Laird, P.~S. Rosenbloom, and E.~Altmann, ``Formulating the problem space computational model,'' in \emph{The Soar papers (vol. II) research on integrated intelligence}, 1993, pp. 1321--1359.

\bibitem{laird2022analysis}
J.~E. Laird, ``An analysis and comparison of act-r and soar,'' \emph{arXiv preprint arXiv:2201.09305}, 2022.

\bibitem{zhou2025hybrid}
R.~Zhou, H.~Cao, J.~Huang, X.~Song, J.~Huang, and Z.~Huang, ``Hybrid lane change strategy of autonomous vehicles based on soar cognitive architecture and deep reinforcement learning,'' \emph{Neurocomputing}, vol. 611, p. 128669, 2025.

\bibitem{laird2022introduction}
J.~Laird, ``Introduction to the soar cognitive architecture,'' 2022.

\bibitem{wray2024eliciting}
R.~E. Wray, J.~R. Kirk, and J.~E. Laird, ``Eliciting problem specifications via large language models,'' \emph{arXiv preprint arXiv:2405.12147}, 2024.

\bibitem{Yujia2022Competition}
Y.~Li, D.~Choi, J.~Chung, N.~Kushman, J.~Schrittwieser, R.~Leblond, T.~Eccles, J.~Keeling, F.~Gimeno, A.~D. Lago, T.~Hubert, P.~Choy, C.~de~Masson~d'Autume, I.~Babuschkin, X.~Chen, P.-S. Huang, J.~Welbl, S.~Gowal, A.~Cherepanov, J.~Molloy, D.~J. Mankowitz, E.~S. Robson, P.~Kohli, N.~de~Freitas, K.~Kavukcuoglu, and O.~Vinyals, ``Competition-level code generation with alphacode,'' \emph{Science (New York, N.Y.)}, pp. 1092--1097, 2022.

\bibitem{Jiang2024from}
W.~Jiang, X.~Gao, J.~Zhai, S.~Ma, X.~Zhang, and C.~Shen, ``From effectiveness to efficiency: Uncovering linguistic bias in large language model-based code generation,'' \emph{arXiv preprint arXiv:2406.00602}, 2024.

\bibitem{karpukhin2020dense}
V.~Karpukhin, B.~Oguz, S.~Min, P.~S. Lewis, L.~Wu, S.~Edunov, D.~Chen, and W.-t. Yih, ``Dense passage retrieval for open-domain question answering.'' in \emph{EMNLP (1)}, 2020, pp. 6769--6781.

\bibitem{johnson2019billion}
J.~Johnson, M.~Douze, and H.~J{\'e}gou, ``Billion-scale similarity search with gpus,'' \emph{IEEE Transactions on Big Data}, vol.~7, no.~3, pp. 535--547, 2019.

\bibitem{2023Generating}
J.~Ye, C.~Li, L.~Kong, and T.~Yu, ``Generating data for symbolic language with large language models,'' \emph{arXiv preprint arXiv:2305.13917}, 2023.

\bibitem{2024Exploring}
B.~Huang, X.~Wu, Y.~Zhou, J.~Wu, L.~Feng, R.~Cheng, and K.~C. Tan, ``Exploring the true potential: Evaluating the black-box optimization capability of large language models,'' \emph{arXiv preprint arXiv:2404.06290}, 2024.

\bibitem{achiam2023gpt}
J.~Achiam, S.~Adler, S.~Agarwal, L.~Ahmad, I.~Akkaya, F.~L. Aleman, D.~Almeida, J.~Altenschmidt, S.~Altman, S.~Anadkat \emph{et~al.}, ``Gpt-4 technical report,'' \emph{arXiv preprint arXiv:2303.08774}, 2023.

\bibitem{brown2020language}
T.~Brown, B.~Mann, N.~Ryder, M.~Subbiah, J.~D. Kaplan, P.~Dhariwal, A.~Neelakantan, P.~Shyam, G.~Sastry, A.~Askell \emph{et~al.}, ``Language models are few-shot learners,'' \emph{Advances in neural information processing systems}, vol.~33, pp. 1877--1901, 2020.

\bibitem{vaswani2017attention}
A.~Vaswani, N.~Shazeer, N.~Parmar, J.~Uszkoreit, L.~Jones, A.~N. Gomez, {\L}.~Kaiser, and I.~Polosukhin, ``Attention is all you need,'' \emph{Advances in neural information processing systems}, vol.~30, 2017.

\bibitem{sumers2023cognitive}
T.~Sumers, S.~Yao, K.~Narasimhan, and T.~Griffiths, ``Cognitive architectures for language agents,'' \emph{Transactions on Machine Learning Research}, 2023.

\bibitem{yao2023react}
S.~Yao, J.~Zhao, D.~Yu, N.~Du, I.~Shafran, K.~Narasimhan, and Y.~Cao, ``React: Synergizing reasoning and acting in language models,'' in \emph{International Conference on Learning Representations (ICLR)}, 2023.

\bibitem{liu2025large}
J.~Liu, W.~Hao, K.~Cheng, and D.~Jin, ``Large language model-based planning agent with generative memory strengthens performance in textualized world,'' \emph{Engineering Applications of Artificial Intelligence}, vol. 148, p. 110319, 2025.

\bibitem{sharma2025novel}
A.~Sharma, A.~Balasundaram, A.~Shaik, and C.~A. Vaithilingam, ``A novel voice in head actor critic reinforcement learning with human feedback framework for enhanced robot navigation,'' \emph{Scientific Reports}, vol.~15, no.~1, p. 7237, 2025.

\bibitem{zargarzadeh2025decision}
S.~Zargarzadeh, M.~Mirzaei, Y.~Ou, and M.~Tavakoli, ``From decision to action in surgical autonomy: Multi-modal large language models for robot-assisted blood suction,'' \emph{IEEE Robotics and Automation Letters}, 2025.

\bibitem{wu2024cognitive}
S.~Wu, A.~Oltramari, J.~Francis, C.~L. Giles, and F.~E. Ritter, ``Cognitive llms: Towards integrating cognitive architectures and large language models for manufacturing decision-making,'' \emph{arXiv preprint arXiv:2408.09176}, 2024.

\bibitem{jia2016data}
R.~Jia and P.~Liang, ``Data recombination for neural semantic parsing,'' \emph{arXiv preprint arXiv:1606.03622}, 2016.

\bibitem{andreas2019good}
J.~Andreas, ``Good-enough compositional data augmentation,'' \emph{arXiv preprint arXiv:1904.09545}, 2019.

\bibitem{zhong2020grounded}
V.~Zhong, M.~Lewis, S.~I. Wang, and L.~Zettlemoyer, ``Grounded adaptation for zero-shot executable semantic parsing,'' \emph{arXiv preprint arXiv:2009.07396}, 2020.

\bibitem{guo2021revisiting}
Y.~Guo, H.~Zhu, Z.~Lin, B.~Chen, J.-G. Lou, and D.~Zhang, ``Revisiting iterative back-translation from the perspective of compositional generalization,'' in \emph{Proceedings of the AAAI Conference on Artificial Intelligence}, vol.~35, no.~9, 2021, pp. 7601--7609.

\bibitem{wang2021codet5}
Y.~Wang, W.~Wang, S.~Joty, and S.~C. Hoi, ``Codet5: Identifier-aware unified pre-trained encoder-decoder models for code understanding and generation,'' \emph{arXiv preprint arXiv:2109.00859}, 2021.

\bibitem{yang2022addressing}
K.~Yang, O.~Deng, C.~Chen, R.~Shin, S.~Roy, and B.~Van~Durme, ``Addressing resource and privacy constraints in semantic parsing through data augmentation,'' \emph{arXiv preprint arXiv:2205.08675}, 2022.

\bibitem{rosenbaum2022clasp}
A.~Rosenbaum, S.~Soltan, W.~Hamza, A.~Saffari, M.~Damonte, and I.~Groves, ``Clasp: Few-shot cross-lingual data augmentation for semantic parsing,'' \emph{arXiv preprint arXiv:2210.07074}, 2022.

\end{thebibliography}

\clearpage
\appendices
\onecolumn

\section{Basic Knowledge Base}
\label{knowledge}
The Basic Knowledge Base consists of 13 knowledge modules spanning various aspects from syntax to execution mechanisms. Each module includes a collection of fine-grained knowledge points/elements, all of which are described in detail~\ref{tab:knowledge modules}.

\begin{table*}[h!]
  \centering
  \footnotesize
  \renewcommand{\arraystretch}{1.6} 
  \caption{Overview of knowledge modules and points/elements}
   \begin{tabular}{c|p{0.3\textwidth}|p{0.55\textwidth}} 
    \toprule[1.2pt]
    \textbf{No.} & \textbf{Knowledge Module} & \textbf{Knowledge points/elements} \\
    \midrule
    \midrule 
    1 & SOAR Programming Basics & SOAR Productions/Rules, Working Memory Elements (WMEs) structure, Basic rule syntax \& graph-based memory, RHS Functions, Fundamental CLI interactions, \emph{e.g.} \\
    \arrayrulecolor{lightgray}
    \midrule
    \arrayrulecolor{black}
    2 & Input Link \& Data Handling & SOAR dot-notation syntax, Input-link structure, Handling multiple rule instantiations for input, Attribute variables in conditions, \emph{e.g.} \\
    \arrayrulecolor{lightgray}
    \midrule
    \arrayrulecolor{black}
    3 & Operator Fundamentals & Decision Cycle, Operator structure (Proposed/Selected), Rule Types (Proposal, Basic Preference, Apply), \emph{e.g.} \\
    \arrayrulecolor{lightgray}
    \midrule
    \arrayrulecolor{black}
    4 & Output Management & Output-link structure, Unary preferences in output operator proposals, Control RHS functions (interrupt, halt), Designing output operators, \emph{e.g.} \\
    \arrayrulecolor{lightgray}
    \midrule
    \arrayrulecolor{black}
    5 & Multi-Apply Rules \& Data Structures & Implementing linked-list data structures in WM, init operator usage, require preference for initialization, Math RHS Functions, \emph{e.g.} \\
    \arrayrulecolor{lightgray}
    \midrule
    \arrayrulecolor{black}
    6 & Substates \& Hierarchical Problem Solving & Theory of Hierarchical Problem Solving via impasses, Substate creation, State management, Impasse resolution strategies within substates, \emph{e.g.} \\
    \arrayrulecolor{lightgray}
    \midrule
    \arrayrulecolor{black}
    7 & Advanced Debugging \& Complex Conditions & Practical debugging of substate operator logic, Implementing \& debugging condition conjunctions/disjunctions, Efficient WMEs copying strategies between states, \emph{e.g.} \\
    \arrayrulecolor{lightgray}
    \midrule
    \arrayrulecolor{black}
    8 & Code Organization \& Advanced Techniques & Best practices for SOAR project structure \& file organization, Error catching mechanisms, Parallel summation strategies, Negated conjunctions for complex tests, Documentation standards, \emph{e.g.} \\
    \arrayrulecolor{lightgray}
    \midrule
    \arrayrulecolor{black}
    9 & Advanced Preferences \& Filtering & In-depth preference combinations, reject preferences, Random tie-breaking using indifferent preferences, \emph{e.g.} \\
    \arrayrulecolor{lightgray}
    \midrule
    \arrayrulecolor{black}
    10 & SOAR Markup Language (SML) Basics & SML API vs. SML messaging, SOAR Kernels and Agent management via SML, SML classes for WME \& Identifier interaction, Basic SML exception handling, \emph{e.g.} \\
    \arrayrulecolor{lightgray}
    \midrule
    \arrayrulecolor{black}
    11 & SML Events \& Dynamic Interaction & Utilizing SML Event Listeners, Enabling dynamic agent interaction through event-driven programming, Implementing event handlers for deterministic behavioral flow, Managing I/O with SML events, \emph{e.g.} \\
    \arrayrulecolor{lightgray}
    \midrule
    \arrayrulecolor{black}
    12 & SOAR 9.6 Reference Guide & Comprehensive quick reference for SOAR 9.6: Detailed Rule Syntax, WMEs types, Memory Systems (SMEM, EPMEM), Reinforcement Learning commands \& parameters, Full Preference system, Impasse State Structures, Extensive RHS Functions, CLI Commands, \emph{e.g.} \\
    \arrayrulecolor{lightgray}
    \midrule
    \arrayrulecolor{black}
    13 & Visual SOAR IDE Usage & Fundamentals of the Visual SOAR IDE, Project structure \& organization within Visual SOAR, Datamap creation and validation for WMEs, Operator hierarchy management tools, Debugger interface integration, Rule editing features (syntax highlighting, auto-completion), \emph{e.g.} \\
    \bottomrule[1.2pt]
  \end{tabular}
  \label{tab:knowledge modules}
\end{table*}

\clearpage
\section{Dataset \& LLMs}
\label{dataset}
\subsection{Water Jug Problem Dataset}

\begin{table*}[ht!]
  \centering
  \caption{Distribution and illustrative cases of the Water Jug Problem Dataset.}
  \label{tab:case_distribution}
  \renewcommand{\arraystretch}{1.8}
  \small
  \resizebox{0.98\textwidth}{!}{%
  \captionsetup{justification=justified,singlelinecheck=false}%
\begin{tabular}{
    >{\raggedright\arraybackslash}p{3cm}|
    >{\centering\arraybackslash}p{1.2cm}|
    >{\raggedright\arraybackslash}p{11cm}
}
    \toprule[1.2pt]
    \textbf{Category} & \textbf{Quantity} & \textbf{Cases} \\ 
    \midrule
    \midrule
    Easy \scriptsize{(\(Min.DC \le 5\))} & 25 & \mbox{(2,5 \(\rightarrow\) 1)/5}, \mbox{(2,7 \(\rightarrow\) 3)/5}, \mbox{(3,5 \(\rightarrow\) 1)/5}, \mbox{(3,7 \(\rightarrow\) 1)/5}, \mbox{(3,8 \(\rightarrow\) 2)/5}, \mbox{(4,7 \(\rightarrow\) 1)/5}, \mbox{(4,9 \(\rightarrow\) 1)/5}, \mbox{(4,10 \(\rightarrow\) 2)/5}, \mbox{(4,11 \(\rightarrow\) 3)/5}, \mbox{(5,9 \(\rightarrow\) 1)/5}, \mbox{(5,11 \(\rightarrow\) 1)/5}, \mbox{(5,8 \(\rightarrow\) 2)/5}, \mbox{(5,12 \(\rightarrow\) 2)/5}, \mbox{(5,7 \(\rightarrow\) 3)/5}, \mbox{(5,13 \(\rightarrow\) 3)/5}, \mbox{(5,14 \(\rightarrow\) 4)/5}, \mbox{(6,11 \(\rightarrow\) 1)/5}, \mbox{(6,13 \(\rightarrow\) 1)/5}, \mbox{(6,10 \(\rightarrow\) 2)/5}, \mbox{(6,14 \(\rightarrow\) 2)/5}, \mbox{(6,15 \(\rightarrow\) 3)/5}, \mbox{(7,13 \(\rightarrow\) 1)/5}, \mbox{(7,12 \(\rightarrow\) 2)/5}, \mbox{(7,11 \(\rightarrow\) 3)/5}, \mbox{(2,9 \(\rightarrow\) 5)/5} \\
    \arrayrulecolor{lightgray}
    \midrule
    \arrayrulecolor{black}
     Med. \scriptsize{(\(5 < Min.DC \le 10\))} & 25 & \mbox{(3,10 \(\rightarrow\) 1)/7}, \mbox{(3,7 \(\rightarrow\) 2)/7}, \mbox{(3,11 \(\rightarrow\) 2)/7}, \mbox{(4,11 \(\rightarrow\) 1)/7}, \mbox{(4,13 \(\rightarrow\) 1)/7}, \mbox{(4,14 \(\rightarrow\) 2)/7}, \mbox{(4,9 \(\rightarrow\) 3)/7}, \mbox{(4,15 \(\rightarrow\) 3)/7}, \mbox{(5,14 \(\rightarrow\) 1)/7}, \mbox{(5,13 \(\rightarrow\) 2)/7}, \mbox{(5,12 \(\rightarrow\) 3)/7}, \mbox{(5,11 \(\rightarrow\) 4)/7}, \mbox{(6,7 \(\rightarrow\) 2)/7}, \mbox{(6,14 \(\rightarrow\) 4)/7}, \mbox{(4,7 \(\rightarrow\) 2)/9}, \mbox{(4,13 \(\rightarrow\) 3)/9}, \mbox{(5,7 \(\rightarrow\) 1)/9}, \mbox{(5,8 \(\rightarrow\) 1)/9}, \mbox{(5,9 \(\rightarrow\) 3)/9}, \mbox{(6,7 \(\rightarrow\) 4)/9}, \mbox{(6,11 \(\rightarrow\) 4)/9}, \mbox{(7,10 \(\rightarrow\) 1)/9}, \mbox{(7,11 \(\rightarrow\) 1)/9}, \mbox{(7,9 \(\rightarrow\) 3)/9}, \mbox{(7,12 \(\rightarrow\) 3)/9} \\
    \arrayrulecolor{lightgray}
    \midrule
    \arrayrulecolor{black}
    Hard \scriptsize{(\(Min.DC > 10\))} & 30 & \mbox{(3,14 \(\rightarrow\) 1)/11}, \mbox{(3,14 \(\rightarrow\) 2)/11}, \mbox{(4,9 \(\rightarrow\) 2)/11}, \mbox{(4,17 \(\rightarrow\) 3)/11}, \mbox{(5,9 \(\rightarrow\) 2)/11}, \mbox{(5,11 \(\rightarrow\) 2)/11}, \mbox{(5,8 \(\rightarrow\) 4)/11}, \mbox{(5,12 \(\rightarrow\) 4)/11}, \mbox{(6,11 \(\rightarrow\) 2)/11}, \mbox{(6,13 \(\rightarrow\) 2)/11}, \mbox{(6,7 \(\rightarrow\) 3)/11}, \mbox{(7,13 \(\rightarrow\) 2)/11}, \mbox{(7,15 \(\rightarrow\) 2)/11}, \mbox{(4,11 \(\rightarrow\) 2)/13}, \mbox{(5,12 \(\rightarrow\) 1)/13}, \mbox{(5,13 \(\rightarrow\) 1)/13}, \mbox{(5,11 \(\rightarrow\) 3)/13}, \mbox{(5,14 \(\rightarrow\) 3)/13}, \mbox{(6,13 \(\rightarrow\) 4)/13}, \mbox{(7,9 \(\rightarrow\) 1)/13}, \mbox{(7,10 \(\rightarrow\) 2)/13}, \mbox{(4,13 \(\rightarrow\) 2)/15}, \mbox{(5,14 \(\rightarrow\) 2)/15}, \mbox{(5,16 \(\rightarrow\) 2)/15}, \mbox{(5,13 \(\rightarrow\) 4)/15}, \mbox{(6,11 \(\rightarrow\) 3)/15}, \mbox{(7,12 \(\rightarrow\) 1)/15}, \mbox{(7,11 \(\rightarrow\) 2)/15}, \mbox{(4,15 \(\rightarrow\) 2)/17}, \mbox{(6,13 \(\rightarrow\) 3)/17} \\
    \arrayrulecolor{lightgray}
    \midrule
    \arrayrulecolor{black}
    Variant \scriptsize{($n=3$)} & 20 & \mbox{(3,5,9 \(\rightarrow\) 2)/3}, \mbox{(4,7,9 \(\rightarrow\) 2)/3}, \mbox{(4,9,11 \(\rightarrow\) 2)/3}, \mbox{(4,11,13 \(\rightarrow\) 2)/3}, \mbox{(5,11,13 \(\rightarrow\) 2)/3}, \mbox{(4,7,9 \(\rightarrow\) 3)/3}, \mbox{(3,5,9 \(\rightarrow\) 1)/4}, \mbox{(3,7,11 \(\rightarrow\) 1)/4}, \mbox{(3,8,13 \(\rightarrow\) 2)/4}, \mbox{(2,5,9 \(\rightarrow\) 1)/5}, \mbox{(4,7,9 \(\rightarrow\) 1)/5}, \mbox{(4,9,11 \(\rightarrow\) 1)/5}, \mbox{(5,11,14 \(\rightarrow\) 2)/5}, \mbox{(4,9,11 \(\rightarrow\) 3)/5}, \mbox{(4,11,13 \(\rightarrow\) 3)/5}, \mbox{(5,9,11 \(\rightarrow\) 3)/5}, \mbox{(6,7,11 \(\rightarrow\) 3)/5}, \mbox{(3,8,13 \(\rightarrow\) 1)/7}, \mbox{(4,11,13 \(\rightarrow\) 1)/7}, \mbox{(3,7,11 \(\rightarrow\) 2)/7} \\
    \bottomrule[1.2pt]
  \end{tabular}
  }
\end{table*}

\subsection{Configurations for LLMs}

\begin{table}[htbp]
  \centering
  \small
  \renewcommand{\arraystretch}{1.2} 
  \caption{Experimental configurations for LLMs. }
  \label{tab:llm_settings}
  \setlength{\tabcolsep}{4pt}
  \begin{tabularx}{0.65\columnwidth}{@{}l| l| >{\raggedright\arraybackslash}X |c c@{}}
    \toprule[1.2pt]
    \multirow{2}{*}{\textbf{Experiment}} & \multirow{2}{*}{\textbf{LLMs}} & \multirow{2}{*}{\textbf{Versions}} & \multicolumn{2}{c}{\textbf{Temperature}} \\
    \cmidrule(lr){4-5}
                          &                       &                           & \scriptsize{NGS-G} & \scriptsize{NGS-C} \\
    \midrule
    \midrule
    \multirow{2}{*}{Baseline} & Gemini               & gemini-2.5-pro-exp-03-25 & 0.3           & -- \\                      
                             & Qwen                 & qwen-max-2025-01-25       & 0.7           & -- \\
    \arrayrulecolor{lightgray}
    \midrule
    \arrayrulecolor{black}
    \multirow{2}{*}{NL2GenSym} & Gemini               & gemini-2.5-flash-preview-04-17 & 0.3           & 0.7 \\
                               & Qwen                 & qwen-max-2025-01-25       & 0.7           & 0.9 \\
    \arrayrulecolor{lightgray}
    \midrule
    \arrayrulecolor{black}
    Ablation                & Gemini               & gemini-2.5-pro-exp-03-25 & 0.3           & 0.7 \\
    \bottomrule[1.2pt]
  \end{tabularx}
\end{table}

\clearpage
\section{Representative Case}
\label{Representative}
\begin{figure*}[htbp]
    \centering
    \vspace{1mm}
    \includegraphics[width=0.9\textwidth]{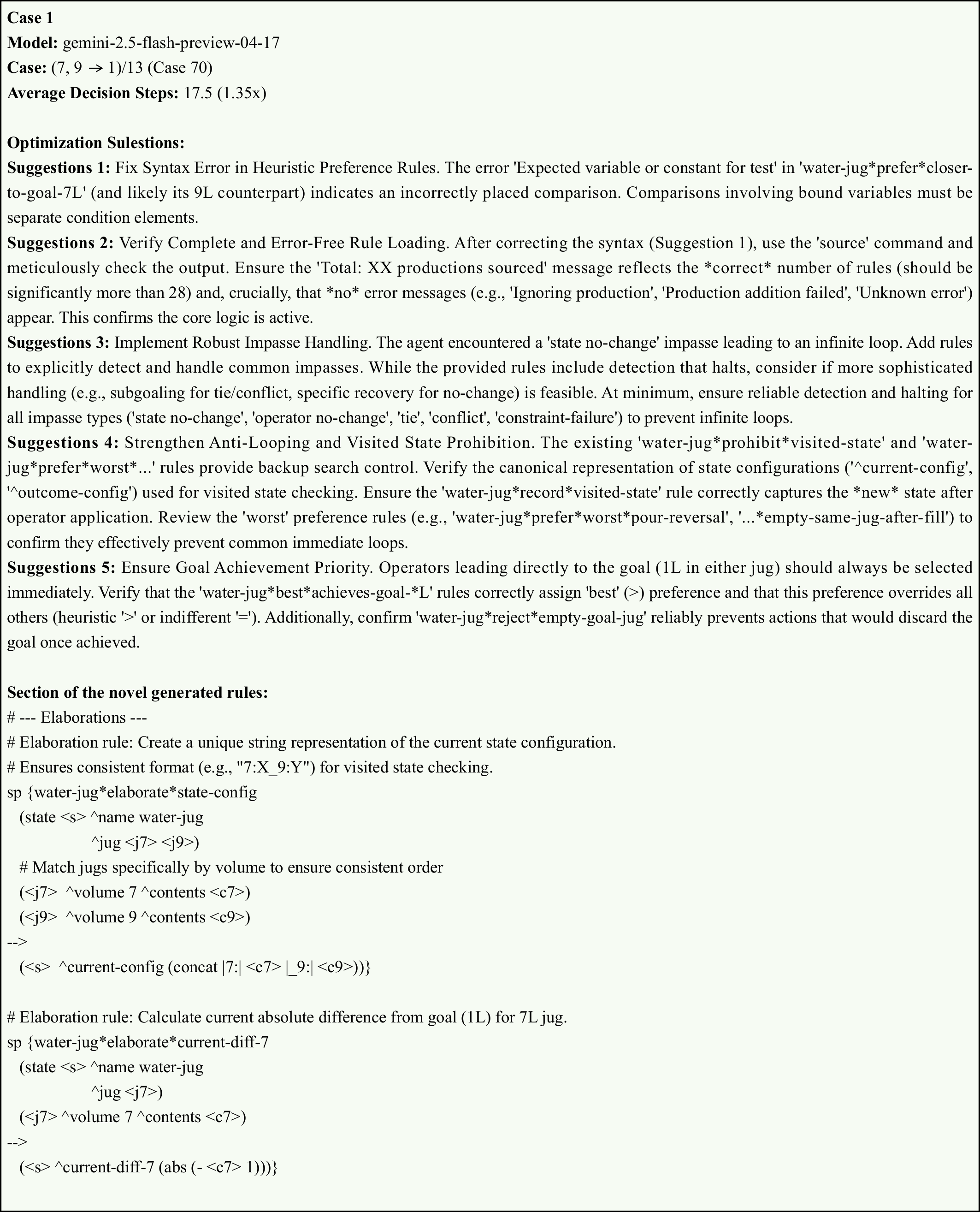}
\end{figure*}
\clearpage
\begin{figure*}[htbp]
    \centering
    \vspace{5mm}
    \includegraphics[width=0.9\textwidth]{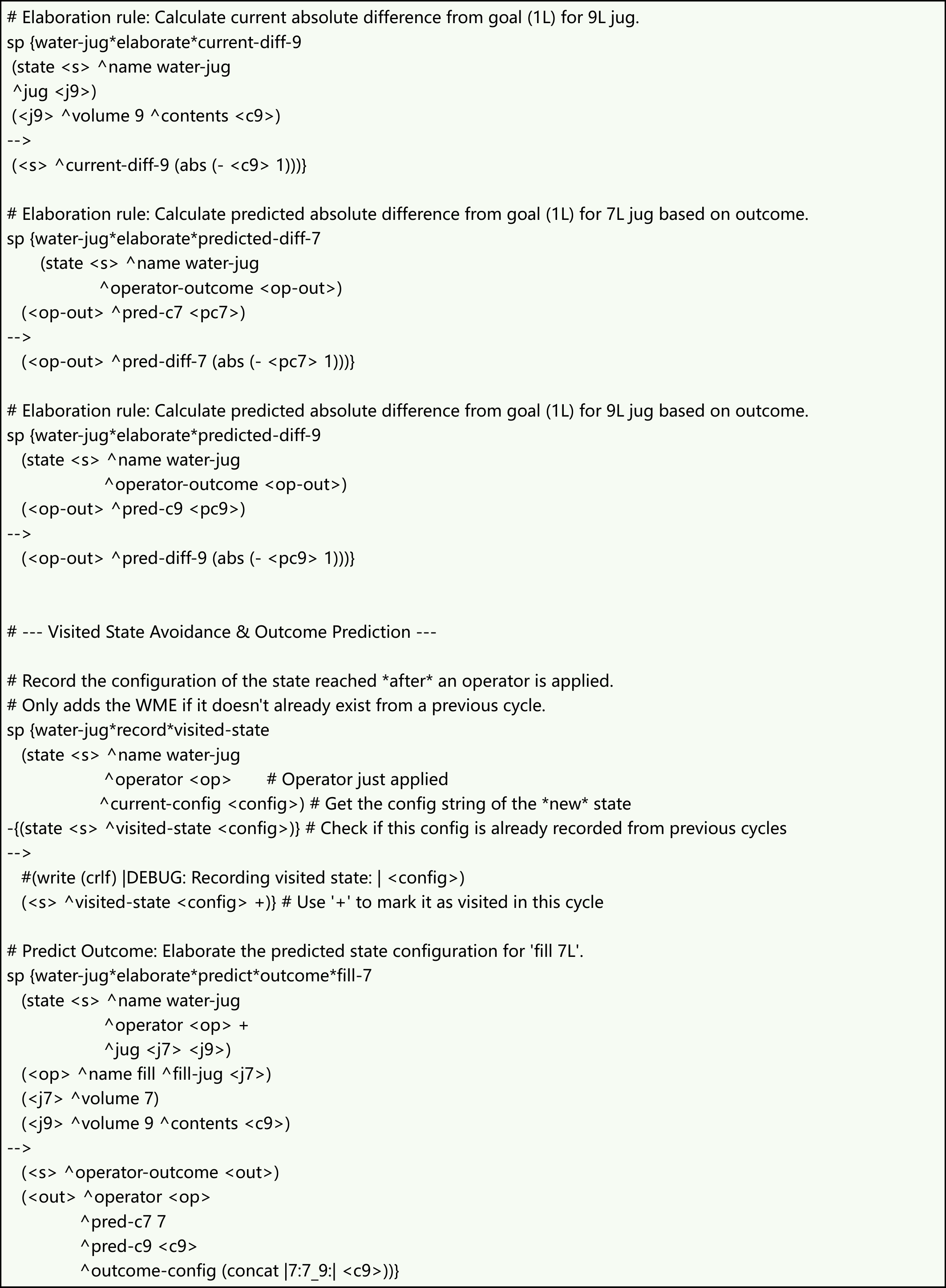}
\end{figure*}
\clearpage
\begin{figure*}[htbp]
    \centering
    \includegraphics[width=0.9\textwidth]{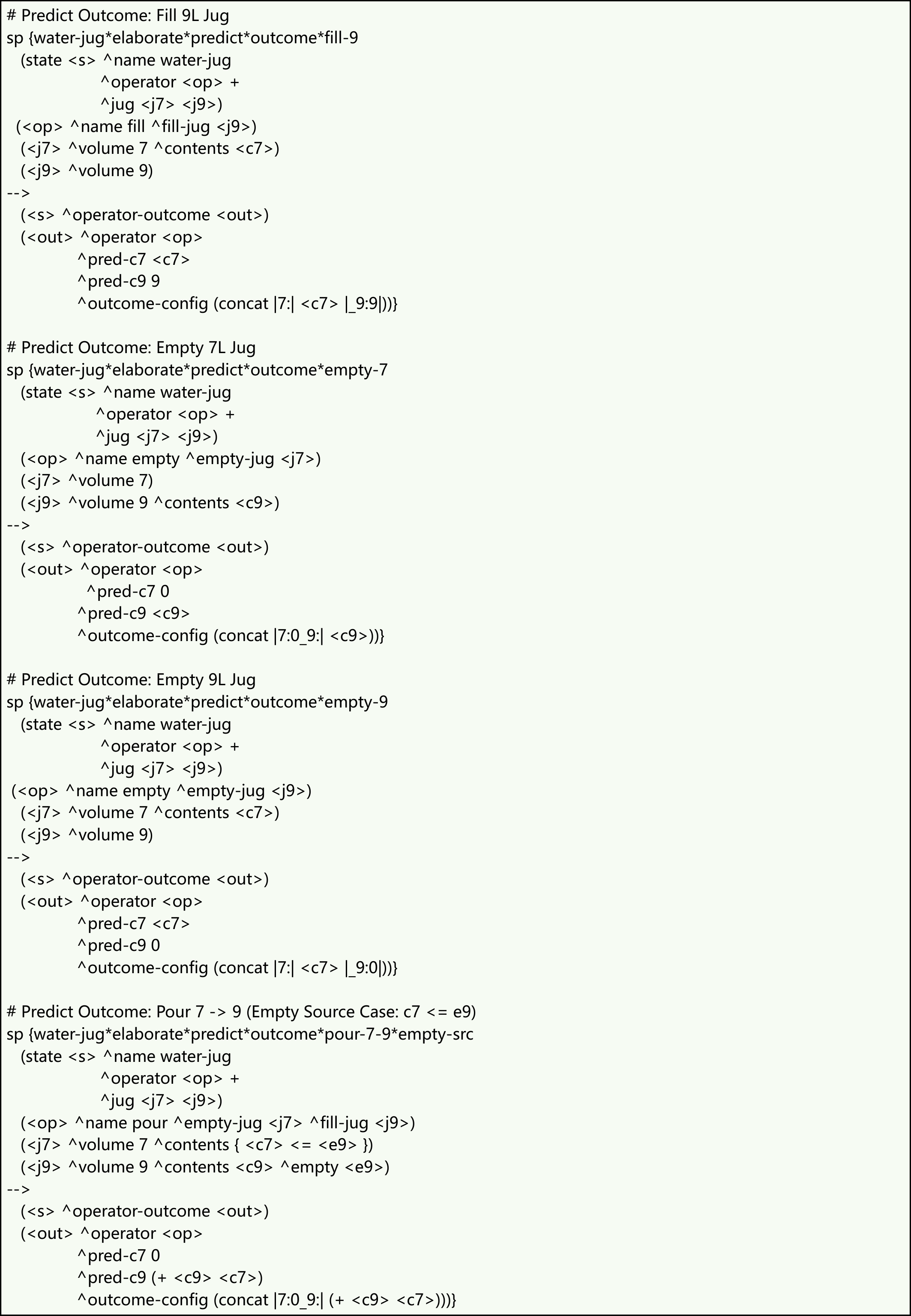}
\end{figure*}
\clearpage
\begin{figure*}[htbp]
    \centering
    \includegraphics[width=0.9\textwidth]{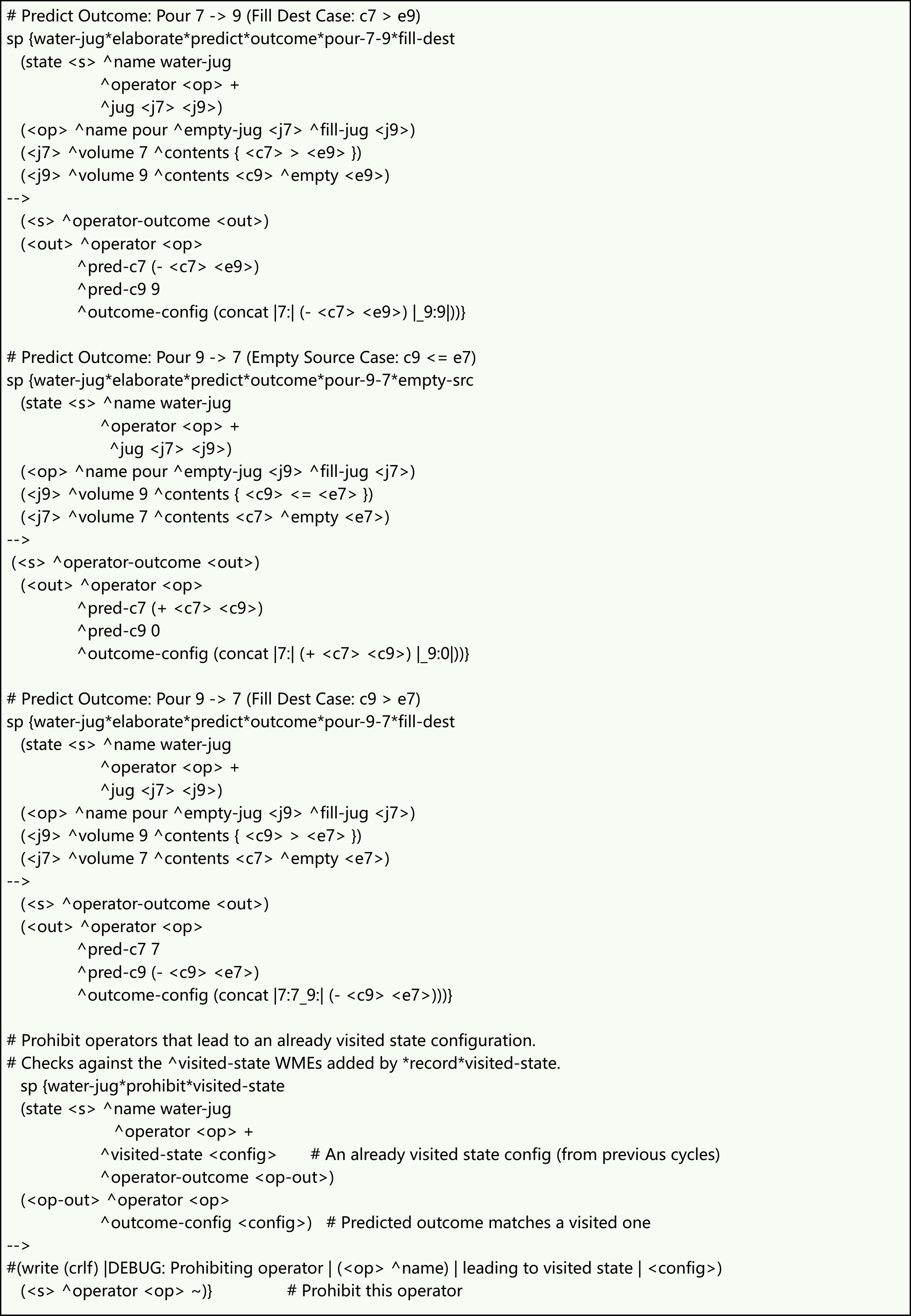}
\end{figure*}
\clearpage
\begin{figure*}[htbp]
    \centering
    \includegraphics[width=0.9\textwidth]{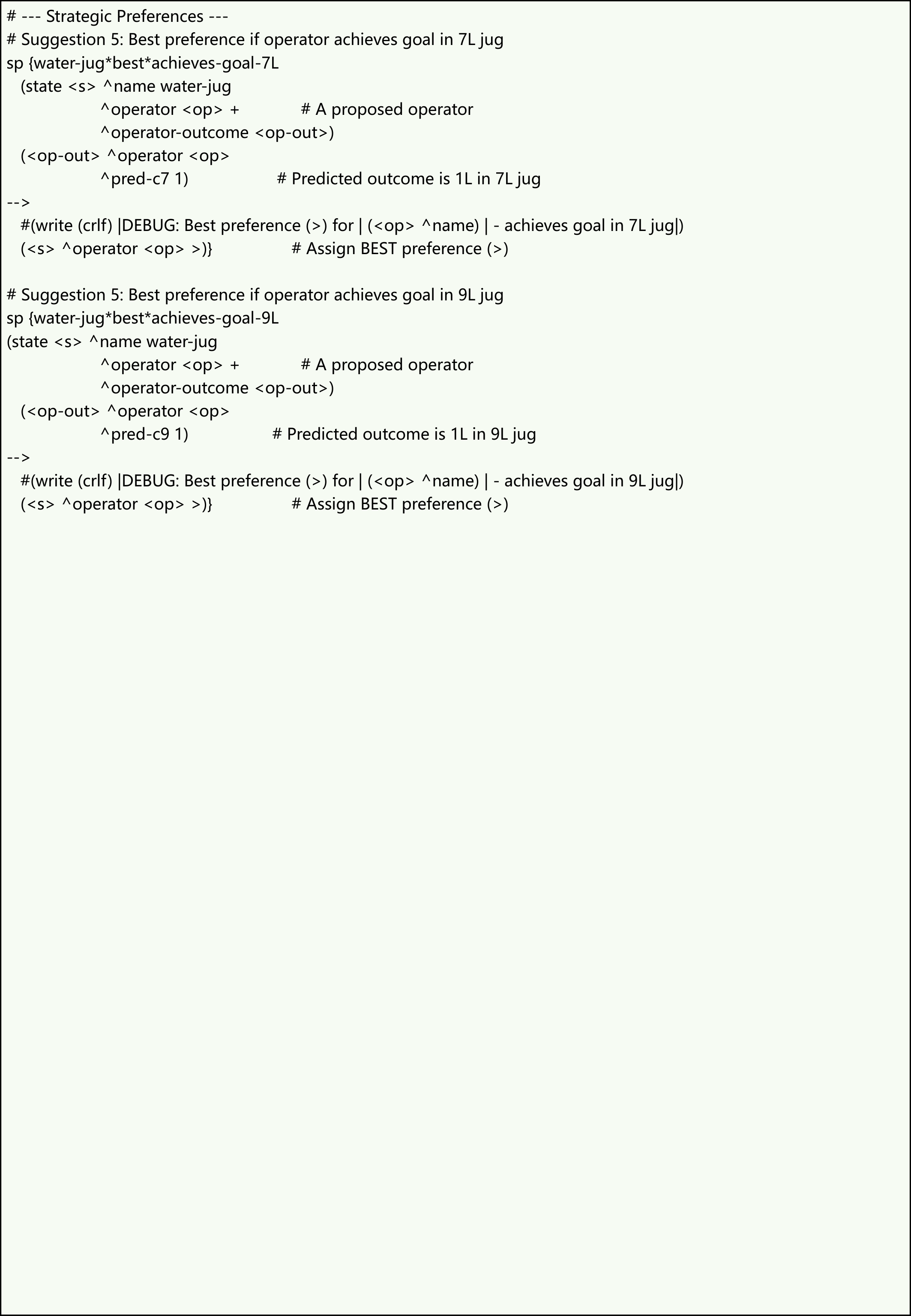}
\end{figure*}
\clearpage
\begin{figure*}[htbp]
    \centering
    \includegraphics[width=0.9\textwidth]{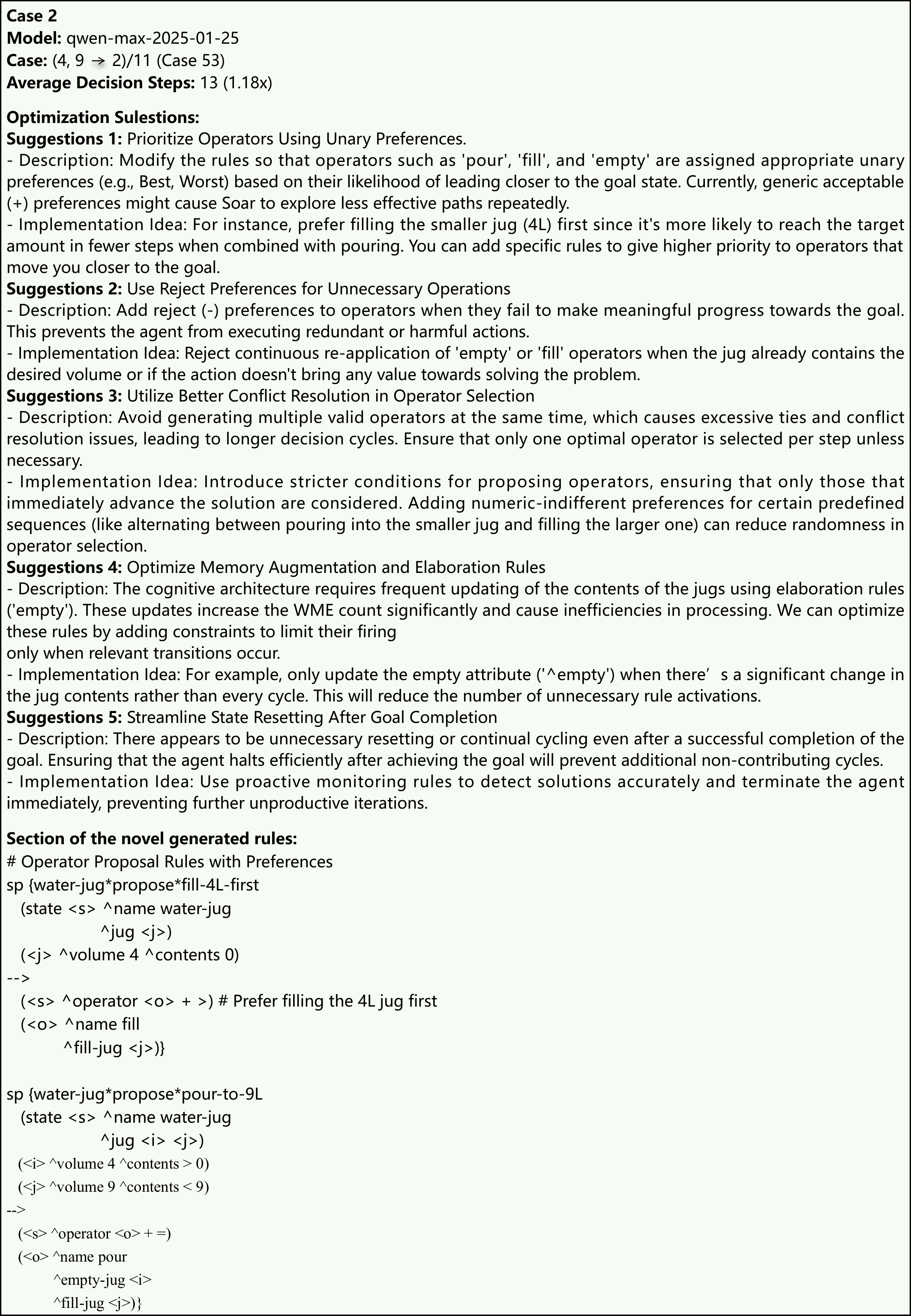}
\end{figure*}
\clearpage
\begin{figure*}[htbp]
    \centering
    \includegraphics[width=0.9\textwidth]{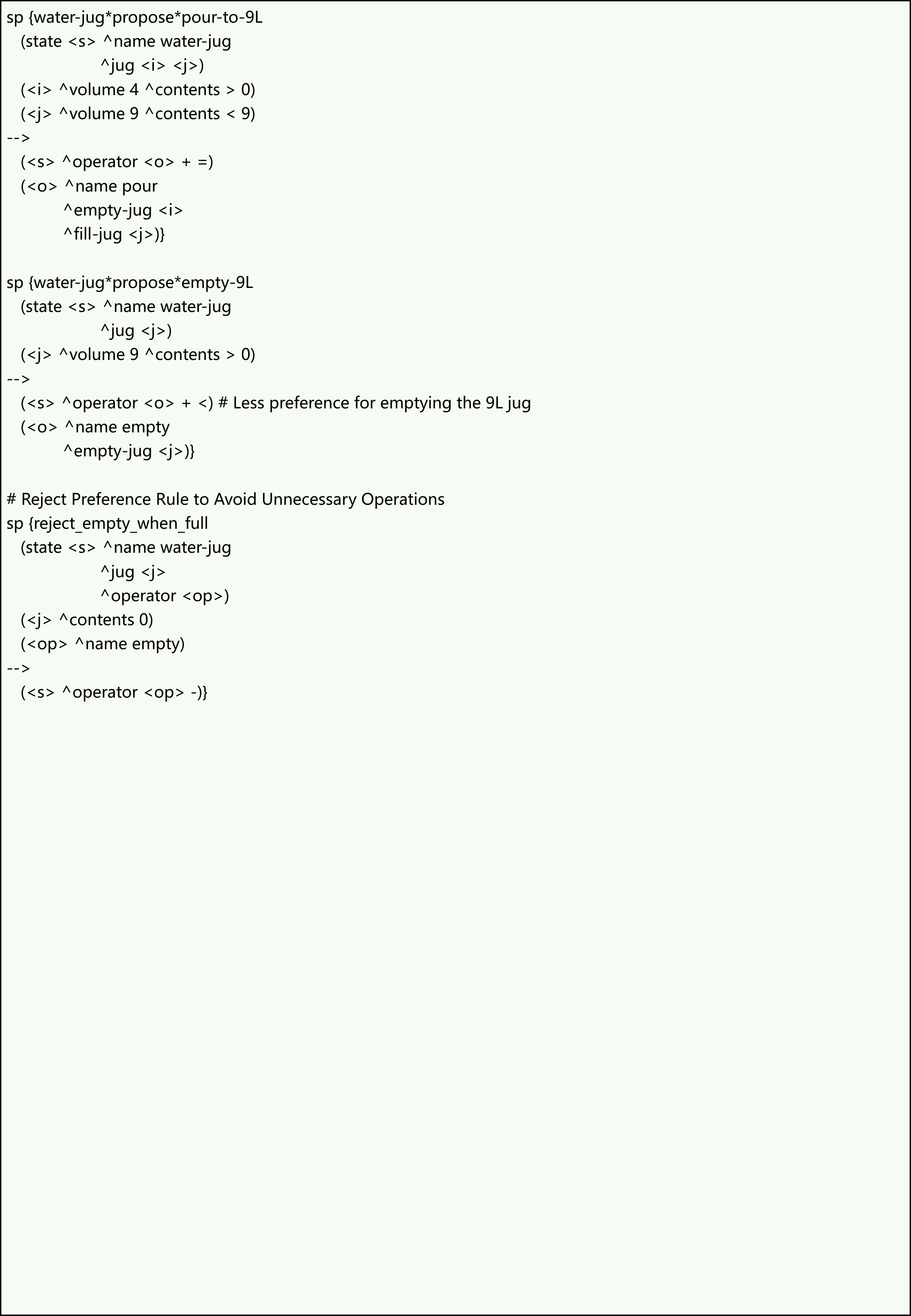}
\end{figure*}

\end{document}